\documentclass[18pt,journal]{article}
\usepackage{amsmath,amsfonts}
\usepackage{algorithmic}
\usepackage{algorithm}
\usepackage{array}
\usepackage{textcomp}
\usepackage{stfloats}
\usepackage{url}
\usepackage{verbatim}
\usepackage{graphicx}
\usepackage{cite}
\usepackage{subcaption}
\usepackage{blindtext} 
\usepackage[a4paper, total={6in, 8in}]{geometry}
\usepackage{graphics}
\usepackage{graphicx}
\usepackage{xcolor}
\usepackage{tikz}
\usepackage{graphicx}
\usepackage{caption}
\usepackage{float}
\usepackage{etoolbox}
\usepackage{graphicx}
\usepackage{mathtools}
\usepackage{amsmath}
\usepackage{commath}

\usepackage{setspace}
\usepackage{multirow}

\usepackage{microtype} 

\usepackage[english]{babel} 

\usepackage{booktabs} 

\usepackage{lettrine} 

\usepackage{enumitem} 

\usetikzlibrary{positioning,calc}

\begin{document}
\baselineskip=0.75cm

\title{Brain-inspired Computational Modeling of Action Recognition with Recurrent Spiking Neural Networks Equipped with Reinforcement Delay Learning}

\author{Alireza Nadafian, Milad Mozafari, Timothée Masquelier, Mohammad Ganjtabesh}

\date{}
\maketitle

\begin{abstract}
\large

\noindent The growing interest in brain-inspired computational models arises from the remarkable problem-solving efficiency of the human brain. Action recognition, a complex task in computational neuroscience, has received significant attention due to both its intricate nature and the brain's exceptional performance in this area. Nevertheless, current solutions for action recognition either exhibit limitations in effectively addressing the problem or lack the necessary biological plausibility. Deep neural networks, for instance, demonstrate acceptable performance but deviate from biological evidence, thereby restricting their suitability for brain-inspired computational studies. On the other hand, the majority of brain-inspired models proposed for action recognition exhibit significantly lower effectiveness compared to deep models and fail to achieve human-level performance. This deficiency can be attributed to their disregard for the underlying mechanisms of the brain. In this article, we present an effective brain-inspired computational model for action recognition. We equip our model with novel biologically plausible mechanisms for spiking neural networks that are crucial for learning spatio-temporal patterns. The key idea behind these new mechanisms is to bridge the gap between the brain's capabilities and action recognition tasks by integrating key biological principles into our computational framework. Furthermore, we evaluate the performance of our model against other models using a benchmark dataset for action recognition, DVS-128 Gesture. The results show that our model outperforms previous biologically plausible models and competes with deep supervised models. 

\end{abstract}

\section{Introduction}

The human brain, renowned for its remarkable capacity, has sparked considerable interests in the development of brain-inspired computational models. These models hold great potential for solving a wide range of complex problems and offer valuable insights into the algorithms and the structures underlying the brain's complex behaviors. Although significant progress has been made in this field, achieving an accurate and comprehensive model for the human brain is still a long way off.

The initial steps towards developing brain-inspired computational models were taken in the 1940s by McCulloch and Pitts \cite{mcculloch}. They scientists presented a logical model of a neuron that functioned based on the summation of input signals to generate an output. The output was created in such a way that if the sum of received inputs exceeded a specified threshold, the neuron fires and sends its output to other neighboring neurons. Although the proposed neural model was very simple, it eventually evolved into the fundamental building block of the most powerful computational models inspired by the brain, namely the neural networks. These networks have became very important over time and continues to expand day by day due to their applications in various fields. These powerful models have not only had a tremendous impact on the progress of artificial intelligence, machine vision, and information processing, but have also found a special place in other areas, including medicine, the environment, and even art.

Although the initial idea of neural networks is derived from the brain's structure, many studies do not utilize the underlying mechanisms in the brain. The reason for this can be attributed to a lack of understanding these mechanisms or a lack of sufficient empirical information about them. On the other hand, the highly efficient functioning of the brain motivates the development of models that are biologically plausible to the greatest extent possible. Providing biologically plausible models can not only lead to higher performance but also contribute to a better understanding of the brain and pave the way for numerous applications in medicine and the treatment of structural or functional brain disorders.

One of the problems in which the brain is highly efficient is action recognition. Numerous studies have been conducted in this field, and many models have been proposed to solve this problem. For example, deep learning-based models such as \cite{1.1,1.2} and spiking neural network models that utilizing the error Backpropagation algorithm \cite{2.1,2.2,2.3} have been introduced. Despite the acceptable performance of these models, their mechanisms are far from the mechanisms underlying the brain, thus not effective in understanding the brain. Furthermore, the complex structure of these models requires lots of computational resources and more data for training. However, we know that activity recognition in the brain is performed with minimal energy consumption and in a very short period of time. Therefore, it is of great importance to present a brain-inspired and biologically plausible model that is efficient in solving the action recognition problem.

Recently, research has been conducted to develop brain-inspired and biologically plausible computational models for activity recognition \cite{3.1,3.2}. Although these models are inspired by the brain, they exhibit very poor performance in activity recognition. One of the main factors contributing to this poor performance is the neglection of many underlying mechanisms of the brain. For example, no mechanisms for learning synaptic delays have been explained in these studies, and considering the important role of synaptic delays in learning spatio-temporal patterns, their absence is noticeable. Therefore, in addition to the model's fidelity to brain mechanisms, its efficiency and problem-solving capabilities are of utmost importance. 

In this article we present a computational model based on a spiking neural network (SNN) that exclusively utilizes brain-inspired and biologically plausible mechanisms. We enhance the SNN with innovative bio-plausible mechanisms and demonstrate its effectiveness in solving the action recognition problem. Additionally, we conduct a comparative analysis of our model's performance against state-of-the-art models using the benchmark dataset DVS-128 Gesture. The results reveal that our model outperforms all the existing bio-plausible models and competes with deep neural networks. Furthermore, we delve into the significance of each proposed mechanism and highlight their individual importance in enhancing the model's performance.

\section{Materials and Methods}

\subsection{Reward-modulated Delay Learning}
The basis of learning in the brain is either unsupervised or reinforced, as provided in many biological studies, and to our knowledge supervised learning, such as error backpropagation has no root in biology. The myelination process is one of the crucial learning mechanisms in the brain for comprehending temporal information by adjusting the conduction delay of connected neurons. Recently, a bio-plausible unsupervised delay learning rule (UDL) for tuning the synaptic delays in Spiking Neural Networks has been proposed \cite{delay_learning}. While the UDL approach, combined with Spike-Timing-Dependent Plasticity (STDP), provides an effective way in acquiring spatio-temporal patterns, it fails in distinguishing between various complex spatio-temporal patterns. This limitation arises from the fact that UDL focuses solely on learning the most frequent elements within each pattern, without incorporating feedback from the environment. Consequently, this learning rule does not demonstrate optimal performance in classification or decision-related tasks, as the frequent elements of patterns may not necessarily serve as distinguishing components. To solve this issue, we propose a reward-modulated delay learning rule (RDL).

Biological evidence shows that reward-related neurotransmitters can control the process of myelination and demyelination \cite{rl_myelin}. For example, an increase in the reward neurotransmitters can result in the acceleration of myelination, while the lack of these neurotransmitters can diminish the size of the myelin sheath over the long term. This observation provides a clear understanding of the indispensability of incorporating a reward modality in the delay learning process. The equations for integrating this module into UDL are as follows:

\begin{equation}
\begin{gathered}
    \Delta t_{i,j} = t_{j} - t_{i}- d_{i,j} - \varepsilon,\text{ and} \\
 \Delta d_{i,j} = G(\Delta t_{i,j}) = \begin{cases}
 r(-B_n \exp{\frac{-\Delta t_{i,j} }{\sigma_n}}) & \text{ $\Delta t_{i,j}  \geq 0$}, \\ 
 r(B_p \exp{\frac{\Delta t_{i,j} }{\sigma_p}}) & \text{ $\Delta t_{i,j}  < 0$}, 
 \end{cases}
\end{gathered}
\label{eq:delay_rstdp}
\end{equation}

 where $\Delta t_{i,j}$ is the firing time difference between the pre-synaptic neuron $i$ and post-synaptic neuron $j$ with respect to the current synaptic delays $d_{i,j}$, and $\Delta d_{i,j}$ indicates the amount of delay changes. Also, $B_p$ and  $B_n$ are the parameters that control the amount of increasing and decreasing of the synaptic delay, respectively. Moreover, $\sigma_p$ is the time constant of delay learning rule in the case of the pre-synaptic neuron has causal effect on the firing of post-synaptic neuron and $\sigma_n$ is the time constant for the case of independency of the firings of two connected neurons. \cite{delay_learning} Also, $r$ represents the amount of injected reward/punishment to the network. In the case of positive reward, the learning rule modifies synaptic delays to facilitate the acquisition of the presented temporal pattern. Conversely, when reward is negative, representing a form of punishment, the learning rule adjusts synaptic delays to promote the forgetting of the currently learned temporal feature. The rationale behind this targeted learning and forgetting mechanism in RDL lies in the similarity to UDL in terms of absolute value changes in delays, while differing in polarity.

\subsection{Modified RDL and R-STDP for Inhibitory Neurons}
\label{sec:inh_rudl}
The inhibitory neurons play an important role in controlling the activity of different parts of the brain, constituting approximately 20\% of the total population of brain neurons. Additionally, learning the delays and synaptic weights of these neurons can have a significant impact on understanding and recognizing brain activity. Due to the distinct role of these inhibitory neurons compared to excitatory ones, their learning can also be different. In this section, we will discuss the computational models proposed for learning the weights of inhibitory neurons and then provide a rule for adjusting the synaptic delays of inhibitory neurons.

\begin{figure}[H]
        \centering    
        \begin{subfigure}{0.32\textwidth}
         \begin{tikzpicture}[scale=0.65]
          \draw[->] (-2.75,0) -- (2.75,0) node[right] {$\Delta t$};
          \draw[->] (0,-2) -- (0,2) node[above] {$\Delta w$};
          \draw[dashed][scale=0.5,domain=-5:0,smooth,variable=\x] plot ({\x},{-3* 2.7 ^ (\x / 1.5)});
          \draw[dashed][scale=0.5,domain=0:5,smooth,variable=\x] plot ({\x},{3* 2.7 ^ (-\x / 1.5)});
          \draw[fill=white] (0,-1.5) circle (2pt) (0,-1.5) node[anchor=west] {};
          \filldraw[black] (0,1.5) circle (2pt) node[anchor=east] {};
        \end{tikzpicture}
          \caption{STDP}
          \label{fig:inh_stdp_a}
        \end{subfigure}
        \begin{subfigure}{0.32\textwidth}
         \begin{tikzpicture}[scale=0.65]
          \draw[->] (-2.75,0) -- (2.75,0) node[right] {$\Delta t$};
          \draw[->] (0,-2) -- (0,2) node[above] {$\Delta w$};
          \draw[dashed][scale=0.5,domain=-5:0,smooth,variable=\x] plot ({\x},{3* 2.7 ^ (\x / 1.5)});
          \draw[dashed][scale=0.5,domain=0:5,smooth,variable=\x] plot ({\x},{-3* 2.7 ^ (-\x / 1.5)});
          \draw[fill=white] (0,-1.5) circle (2pt) (0,-1.5) node[anchor=west] {};
          \filldraw[black] (0,1.5) circle (2pt) node[anchor=east] {};
        \end{tikzpicture}
         \caption{Reverse STDP}
         \label{fig:inh_stdp_b}
        \end{subfigure}
        \begin{subfigure}{0.32\textwidth}
         \begin{tikzpicture}[scale=0.65]
          \draw[->] (-2.75,0) -- (2.75,0) node[right] {$\Delta t$};
          \draw[->] (0,-2) -- (0,2) node[above] {$\Delta w$};
          \draw[dashed][scale=0.5,domain=-5:0,smooth,variable=\x] plot ({\x},{+3* 2.7 ^ (\x / 1.5)});
          \draw[dashed][scale=0.5,domain=0:5,smooth,variable=\x] plot ({\x},{+3* 2.7 ^ (-\x / 1.5)});
          \filldraw[black] (0,1.5) circle (2pt) node[anchor=east] {};
        \end{tikzpicture}
        \caption{Symmetric STDP}
        \label{fig:inh_stdp_a}
        \end{subfigure}

    \caption{STDP-based learning rules for the weights of inhibitory neurons}
    \label{fig:inh_stdp}

\end{figure}
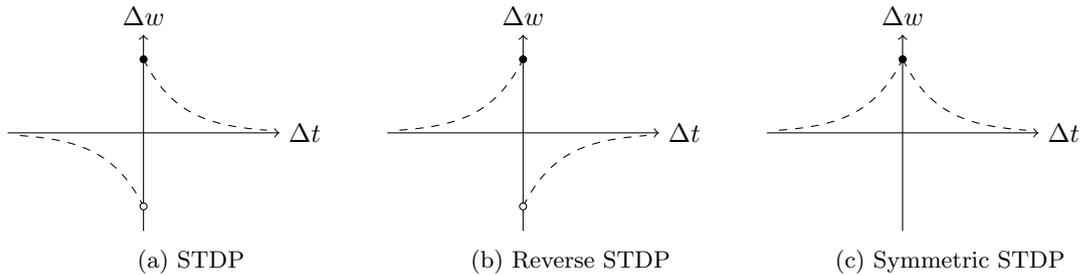

To learn the weights of inhibitory neurons in spiking neural networks, multiple learning rules have been proposed, some of which are illustrated in Figure \ref{fig:inh_stdp} \cite{inh_stdp}. Although there are biological evidences supporting the validity of all these rules, the reinforcement learning rule presented in Equation \ref{eq:inh_rstdp}, which corresponds to spike-timing-dependent reinforcement learning defined for inhibitory neurons, exhibits higher efficiency in activity recognition tasks. Its functionality is such that if a decision leads to a reward in the model, the connections of inhibitory neurons are weakened, resulting in the model moving towards increasing the probability of the chosen decision for the input pattern. Additionally, if the model makes an incorrect decision and is punished accordingly, the connections of inhibitory neurons become stronger, preventing a similar decision from being made upon observing the input pattern again.

\begin{equation}
\begin{gathered}
 \Delta t_{i,j} = t_{j} - t_{i}- d_{i,j}, \\
 \Delta w_{i,j} = F(\Delta t_{i,j}) = \begin{cases}
 r(A_p \exp{\frac{-\Delta t_{i,j} }{\tau_p}}) & \text{ $\Delta t_{i,j}  \geq 0,$} \\ 
 r(-A_n \exp{\frac{\Delta t_{i,j} }{\tau_n}}) & \text{ $\Delta t_{i,j}  < 0.$}
 \end{cases}
\end{gathered}
\label{eq:inh_rstdp}
\end{equation}

Similar to synaptic weights, the synaptic delays of inhibitory neurons can also play an important role in the differentiation and perception of activities.  Equation \ref{eq:inhـrudl} provides the learning rule for adjusting the delays of inhibitory neurons. As a result of the adjustment, if an inhibitory pre-synaptic neuron doesn't have an impact on the activity of the post-synaptic neuron and the network made a correct decision (receiving reward), in addition to the decrease in the strength of the inhibitory synaptic weight by equation \ref{eq:inh_rstdp}, increasing the synaptic delay leads to the loss of inhibitory relation between the pre-synaptic and post-synaptic neurons. So, due to the coherent activity and correct decision of the network, the inhibition is prevented when the input pattern is observed again. Furthermore, in the case of the incorrect decision of the network (receiving punishment), according to the provided learning and in a similar manner, the level of inhibition of the current network activity increases to prevent its incorrect decision in future.

\begin{equation}
\begin{gathered}
   \Delta t_{i,j} = t_{j} - t_{i}- d_{i,j}, \\
 \Delta d_{i,j} = H(\Delta t_{i,j}) = \begin{cases}
 r(B_n \exp{\frac{-\Delta t_{i,j} }{\sigma_n}}) & \text{ $\Delta t_{i,j}  \geq 0$} \\ 
 r(-B_p \exp{\frac{\Delta t_{i,j} }{\sigma_p}}) & \text{ $\Delta t_{i,j}  < 0$} 
 \end{cases}.
\end{gathered}
\label{eq:inhـrudl}
\end{equation}

\subsection{Delay Learning for Non-linear Structures}

\begin{figure}[h]
\centering
\includegraphics[width=\linewidth]{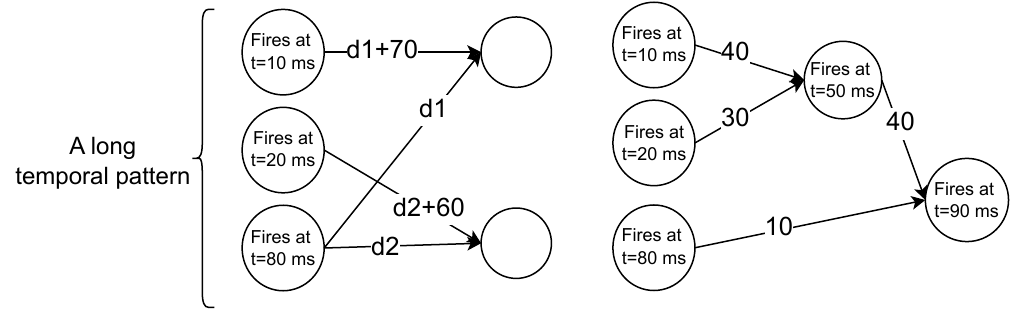}
\caption{Information (spike) flow in layered (left) and unstructured networks (right). }
\label{fig:recurrent_delay}
\end{figure}

Using a layered neural network structure for learning long-term or alternating patterns will not be efficient. However, using a set of neurons with lateral and random connections can address this problem. For example, Figure \ref{fig:recurrent_delay} illustrates how a non-layered structure can help in learning long-term temporal patterns. As seen in this figure, due to the large distance between the firing time of the third and other neurons, in order to receive the message simultaneously with any other neuron, we need a very long synaptic delay, which is not biologically plausible. On the other hand, the existence of a connection between the two postsynaptic neurons retains information related to the firing time of the first two neurons until it combines with the information related to the firing of the third neuron at the appropriate time. Therefore, using a non-layered structure is essential for the perception and recognition of an activity. It was shown that STDP learns meaningful patterns effectively even in an unstructured network \cite{polychron}. So, due to the similarity of RDL and STDP, we can use RDL for the adjustment of the delays of all synapses in an unstructured network (like forward and lateral synapses).

\subsection{Intervallic Homeostasis Regulation on Neural Activity}

An inherent challenge in the homeostasis regulation proposed in \cite{delay_learning} lies in the determination of an optimal value for the target neuron activity, denoted by $R_{target}$. Particularly, in many instances, it remains unclear what specific value would drive the model towards an equilibrium state. To overcome this challenge, we present an alternative approach that considers a range $[R_{min}, R_{max}]$ for target neuron activity instead of a single value as $R_{target}$. Specifically, $R_{min}$ represents the minimum expected level of neuron activity, while $R_{max}$ represents its maximum expected level. By adopting this range-based approach, we aim to offer a more adaptable and flexible solution for determining and regulating neural activity levels in the context of homeostasis. The equations for intervallic homeostasis regulation on neural activity by adjusting weights and synaptic delays can be written as follows:

\begin{equation}
\begin{gathered}
 K = \begin{cases}
   k_{max} \frac{R_{max} - R_{observed}}{R_{max}} & \text {$R_{max} < R_{observed}$,} \\
   k_{min} \frac{R_{observed}-R_{min}}{R_{min}} & \text {$R_{min} > R_{observed}$,} \\
   0 & o.w.
 \end{cases}\\
 d_{new} = d_{current} - \lambda_d \times K,\\
w_{new} = w_{current} + \lambda_w \times K.
\end{gathered}
\label{eq:mod_homeo}
\end{equation}

 The values of $k_{max}$ and $k_{min}$ determine the amount of adjustments in homeostatic learning when the neuron activity exceeds $R_{max}$ or falls below $R_{min}$, respectively. Additionally, when neural activity level remains in the range of $[R_{min}, R_{max}]$, this regulatory mechanism does not apply any changes to the weights and delays of the synapses connected to the target neuron.

\subsection{Intervallic Threshold Adaptation}

The short-term regulation of neuron activity is accomplished through the homeostasis regulation. Nonetheless, in numerous instances, the activity of a neuron can be influenced by its inherent limitations concerning sparse connectivity, synaptic weights and synaptic delays. For instance, in situations where the number of pre-synaptic neurons connected to a post-synaptic neuron is small, the neuron may struggle to reach its desired activation level. Moreover, increasing the weights of the synapses is not a feasible solution, as they eventually reach their maximum limit, resulting in insufficient input for the post-synaptic neuron. Consequently, adjusting the threshold of the neuron becomes necessary in such scenarios. In this manner, if the neuron's activity level in the long term is lower than a specific target, the threshold is decreased, and if it exceeds the target, it is increased. Similar to the homeostasis regulation, setting a specific target value for this adaptive range can lead to imbalance or inconsistent behavior by neurons. Therefore, we consider a valid interval for the target activity of neurons instead of a single value. The equations for the intervallic threshold adaptation can be written as follows:

 \begin{equation}
\begin{gathered}
\Delta \theta = \begin{cases}
   \theta_{+}  & \text {$R_{observed} < R_{min}$,}  \\
   \theta_{-}  & \text {$ R_{observed} > R_{max}$,}  \\
   0 & o.w.
 \end{cases}\\
\end{gathered}
\label{eq:ta_interval}
\end{equation}

where $\theta_{+}$ ($\theta_{-}$) determines the amount of increase (decrease) in the neuron's threshold if its activity level falls below $R_{min}$ (exceeds $R_{max}$).

\subsection{Homeostasis on Decision}

In addition to the regulation of neuronal activity within a spiking neural network (SNN), there arises a need for another mechanism to modulate the decision-making process of the network. Specifically, in the context of a classification problem, this mechanism plays a crucial role in preventing the network from persistently predicting a single class without considering other possibilities. The significance of avoiding such occurrences in the learning process of SNNs warrants further investigation. The bio-plausible learning rules employed in these networks, such as RDL and STDP, heavily rely on the neural activity. Any change in the distribution of this activity can significantly impact the learning trajectory of the network. In the specific context of decision-making within spiking neural networks, inputs associated with a particular decision that elicit a strong response from the network can monopolize its resources. Consequently, the heightened activity levels across all neurons can slow or even disrupt the learning process for other decisions. Therfore, it becomes imperative to establish a mechanism for regulating the decision-making process within the network. Similar to the concept of homeostasis regulation for neural activity, we modify the synaptic weights and delays associated with neurons corresponding to specific decision groups. These adjustments serve to modulate the activity levels of these neurons, and thereby enabling a controlled and balanced decision-making process. The equations for the proposed homeostatic regulation on the decision can be written as follows:

\begin{equation}
\begin{gathered}
 K^{i} = \frac{P^{i}_{target} - P^{i}_{observed}}{P^{i}_{target}}, \\
w^{i}_{new} = w^{i}_{current} - \lambda_w \times K^{i}, \\
d^{i}_{new} = d^{i}_{current} - \lambda_d \times K^{i},
\end{gathered}
\label{eq:decision_homeo}
\end{equation}

In this equation, $P^i_{target}$ and $P^i_{observed}$ represent the target and observed number of occurrences for decision $i$ in the model, respectively. Additionally, $w^i$ and $d^i$ denote the weights and delays associated with neurons corresponding to decision $i$.  In instances where an excessive occurrence of a particular decision is detected, the weights of synapses connected to neurons corresponding to that decision are decreased, and their delays are increased. Consequently, the frequency of that decision being repeated by the model is reduced. Conversely, enhancing the weights and reducing the delays of neurons associated with a decision that is infrequently chosen by the model increases the repetition of that decision.

\subsection{Controlling the Activity of Decision Neurons (Decentralization)}

In the context of action recognition, it is crucial to acknowledge that a particular action can manifest in numerous dissimilar variants. For instance, in the DVS-128 Gesture dataset, the action of clockwise hand movement has various types such as different speeds, hand positions, and the size of the person's hand performing the activity. Consequently, a model's comprehension of an action should not be confined to a singular \textbf{centralized} understanding, but rather encompass the potential for diverse types. Therefore, it is expected from an ideal model to exhibit distinct behavior when receiving various types of a single action. For example, in spiking neural networks it is expected that only a subset of neurons correspond to the specific action become active. This subset of neurons is responsible for learning and representing the particular sub-type of the action.

On the other hand, allocation of sufficient resources for different decisions is one of the encountered challenges in spiking neural networks. Insufficient availability of resources, represented by a limited number of active neurons, hinders the network's ability to learn complex patterns effectively. Conversely, incorporating a very large number of neurons in the network (and using a mechanism like homeostasis to enforce their activity) not only increases the energy consumption, but also leads to many neurons being involved in learning simple patterns. Such an arrangement lacks justification in light of the existing biological evidence highlighting the remarkable efficiency of the brain.

Hence, this question arises: how can we properly utilize the increased number of neurons and network resources to learn all variants of a specific class of action? To address this question, we propose the \textbf{decentralization} in a spiking neural network. This process operates in such a way that, for each input presented to the network, only a limited number of neurons (e.g., $DC_{upper}$) corresponding to an action that have fired earlier than the other neurons associated with the same decision are allowed to become active. As a result, each pattern will be processed only by a limited number of neurons. Moreover, this internal competition among the neurons associated with each decision causes them to be divided into smaller groups, with each group learning a different variant of the action. Furthermore, due to the imposed limitations, it is possible to increase the number of network neurons without sacrificing efficiency and increasing energy consumption.

\section{Results}

In this section, our objective is to utilize the newly introduced mechanisms within a model incorporating a spiking neural network as its foundation. Our aim is to propose a biologically plausible model for action recognition. We begin by presenting the structure of our model and describing its training phase. Subsequently, we evaluate its performance on a benchmark dataset and compare it against other state-of-the-art models. Lastly, we validate the significance of each mechanism within the model and assess their influence on the learning process and overall efficiency of the model.

\subsection{Human Activity Recognition}

In order to assess the efficacy of the proposed model, we employ the DVS128-Gesture dataset \cite{dvs} as our benchmark. This dataset comprises 11 distinct activities executed by 29 individuals, recorded under 3 different illuminations conditions. The activities were captured using cameras equipped with dynamic vision sensors. In our model, we impose a maximum duration of about six seconds for each activity, equivalent to a capture of 200 frames at a frame rate of 33 fps. Also, the eleventh class in this dataset represents activities performed in an arbitrary manner. To avoid potential complexities associated with this class, it is excluded from our analysis. Furthermore, the DVS128-Gesture dataset is composed of two distinct sets: the training dataset and the testing dataset. The training dataset includes activities executed by the first 23 individuals, while the remaining six individuals' activities are employed to assess and evaluate the performance of the models. Our proposed model is trained and evaluated in the same manner.

\subsection{Model Structure}

\begin{figure}
     \centering
     \includegraphics[width=\textwidth]{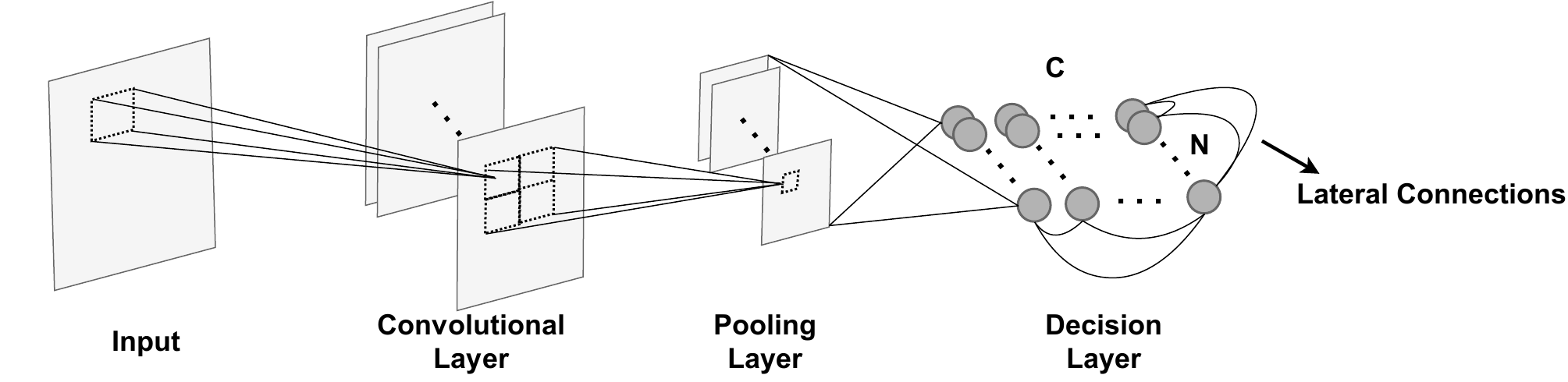}
     \caption{The overall structure of the employed SNN}
     \label{fig:final_model}
\end{figure}

The proposed model is a two-layer spiking neural network featuring leaky integrate-and-fire (LIF) neurons. The overall architecture of the model is depicted in Figure \ref{fig:final_model}. The first layer of the model is a convolutional layer with shared weights and delays, followed by a (spatial) max-pooling module. In this layer, we employ spike-timing-dependent plasticity (STDP) and the unsupervised delay learning rule (UDL) proposed in \cite{delay_learning} to adjust the synaptic weights and delays, respectively. Moreover, intervallic homeostasis regulation is incorporated to regulate the activities of the neurons, within this layer. Consequently, employing these mechanisms would lead this layer to learn local and short repeating spatio-temporal patterns, similar to simple feature maps in the the early layers of the visual cortex. The second layer of the proposed model is the decision layer, which consists of $C\times N$ neurons. Here, $C$ corresponds to the number of output decisions classes, while $N$ indicates the number of neurons associated with each decision group. Notably, each neuron within the decision layer is connected to all neurons in the first layer. Additionally, there exist interconnections among the neurons in the decision layer, leading to recurrent connections and a non-layered structure. This arrangement allows for complex interactions and information flow between neurons, facilitating the decision-making process in the model. In this layer, the learning of synaptic weights is performed using the Reward modulated STDP learning rule proposed in \cite{rstdp_m}, and the learning of delays is achieved through reward-modulated delay learning (RDL). Additionally, if a synapse is connected to an inhibitory neuron, the delay adjustment is done using the rule mentioned in section \ref{sec:inh_rudl}.  Moreover, various mechanisms including intervallic homeostasis regulation on neural activity, homeostasis on decision, intervallic threshold adaptation, and decentralization are employed in this layer to modulate and regulate the network's activity and thus increasing its decision-making capabilities, 

The decision making process is based on a majority voting mechanism that takes into account the spike counts of neurons associated with each potential decision. Moreover, we introduce a reward or punishment signal to the model according to the correct or incorrect decisions, respectively. To this ends, we use a variable, denoted as $r$, to represent the reward or punishment level of the model. We increment (decrement) the value of $r$ by a constant factor for each spike from neurons associated (not associated) with the target decision.

\subsection{Training phase}

\begin{figure}[!h]
     \centering
     \begin{subfigure}[b]{0.45\textwidth}
         \centering
         \includegraphics[width=\textwidth]{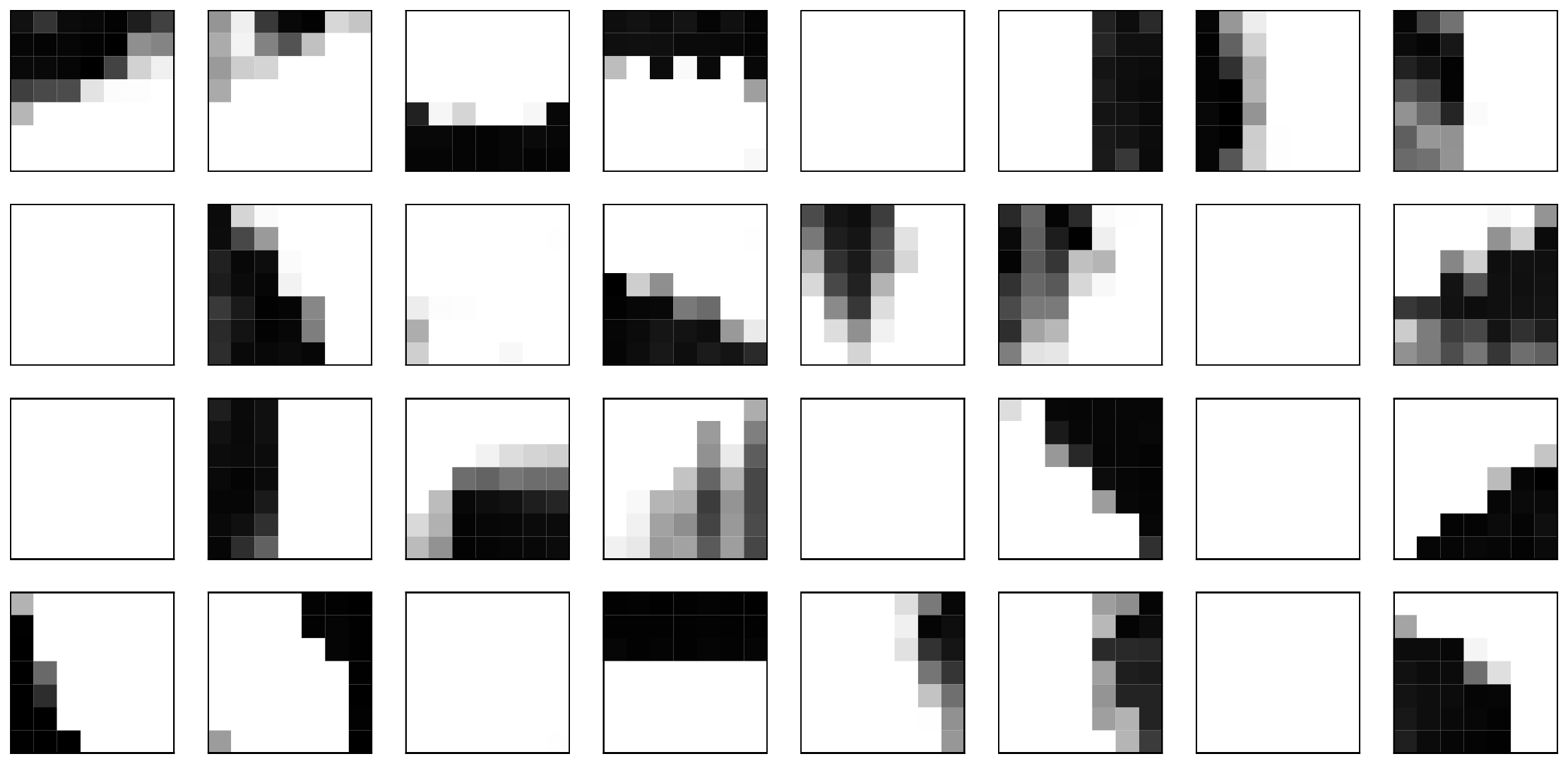}
         \caption{Synaptic weights}
     \end{subfigure}
     \hfill
     \begin{subfigure}[b]{0.45\textwidth}
         \centering
         \includegraphics[width=\textwidth]{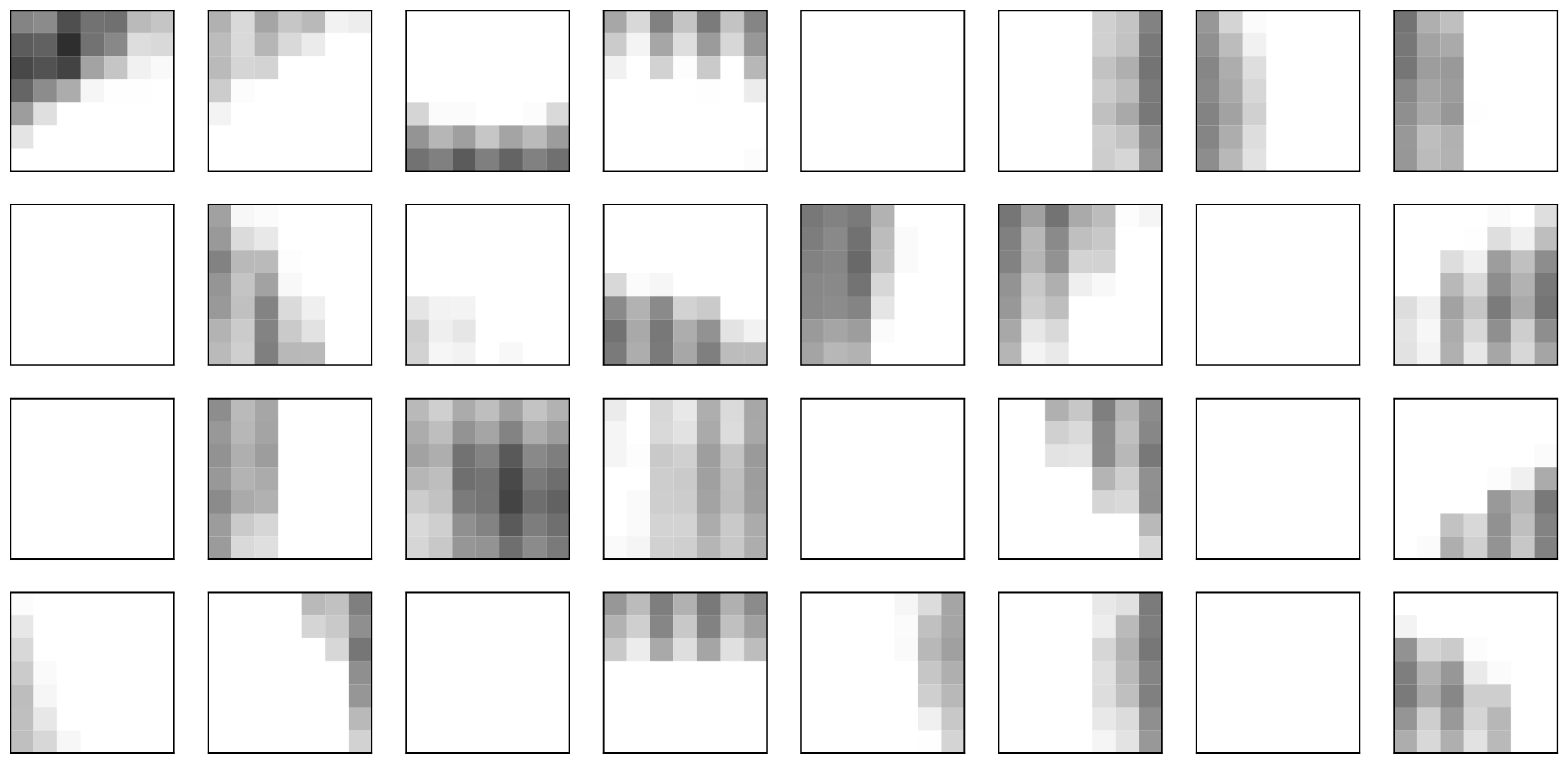}
         \caption{Synaptic delays}
     \end{subfigure}
     \hfill
        \caption{Features learnt in the first layer of the model. Darker color indicates stronger connections (higher weights and lower delays).}
        \label{fig:features_l1}
\end{figure}

According to the structure of the model, the layers are trained sequentially and independently. Initially, the first layer of the model is trained on the train-set in an unsupervised manner, utilizing STDP, UDL, and homeostasis regulation. To do so, we present each action of the dataset to the model one by one. Throughout the presentation of each action, we record the neural activities (spike times) and update the weights and delays after the final spike. We continue training the model until all neurons reach the frozen state according to the delay learning stop condition outlined in \cite{delay_learning}. Note that we don't update the weights and delays of the second layer at this stage. Fig. \ref{fig:features_l1} illustrates the learned features (weights and delays) in the first layer, which primarily consist of simple local movements like a moving line towards a specific direction.  

\begin{figure}
     \centering
     \begin{subfigure}[b]{0.8\textwidth}
         \centering
         \includegraphics[width=\textwidth]{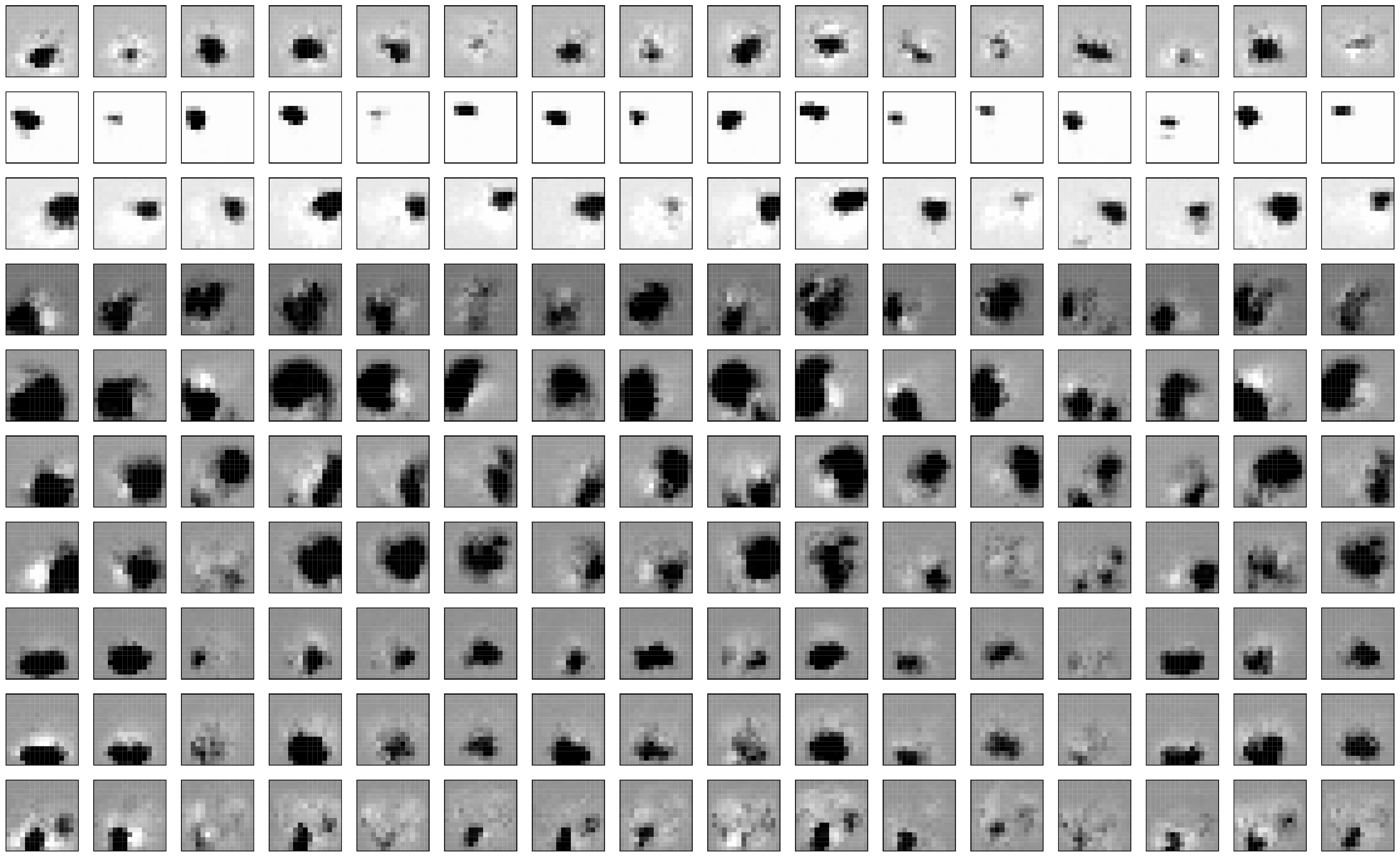}
         \caption{Synaptic weights}
     \end{subfigure}
     \hfill
     \begin{subfigure}[b]{\textwidth}
         \centering
         \includegraphics[width=0.8\textwidth]{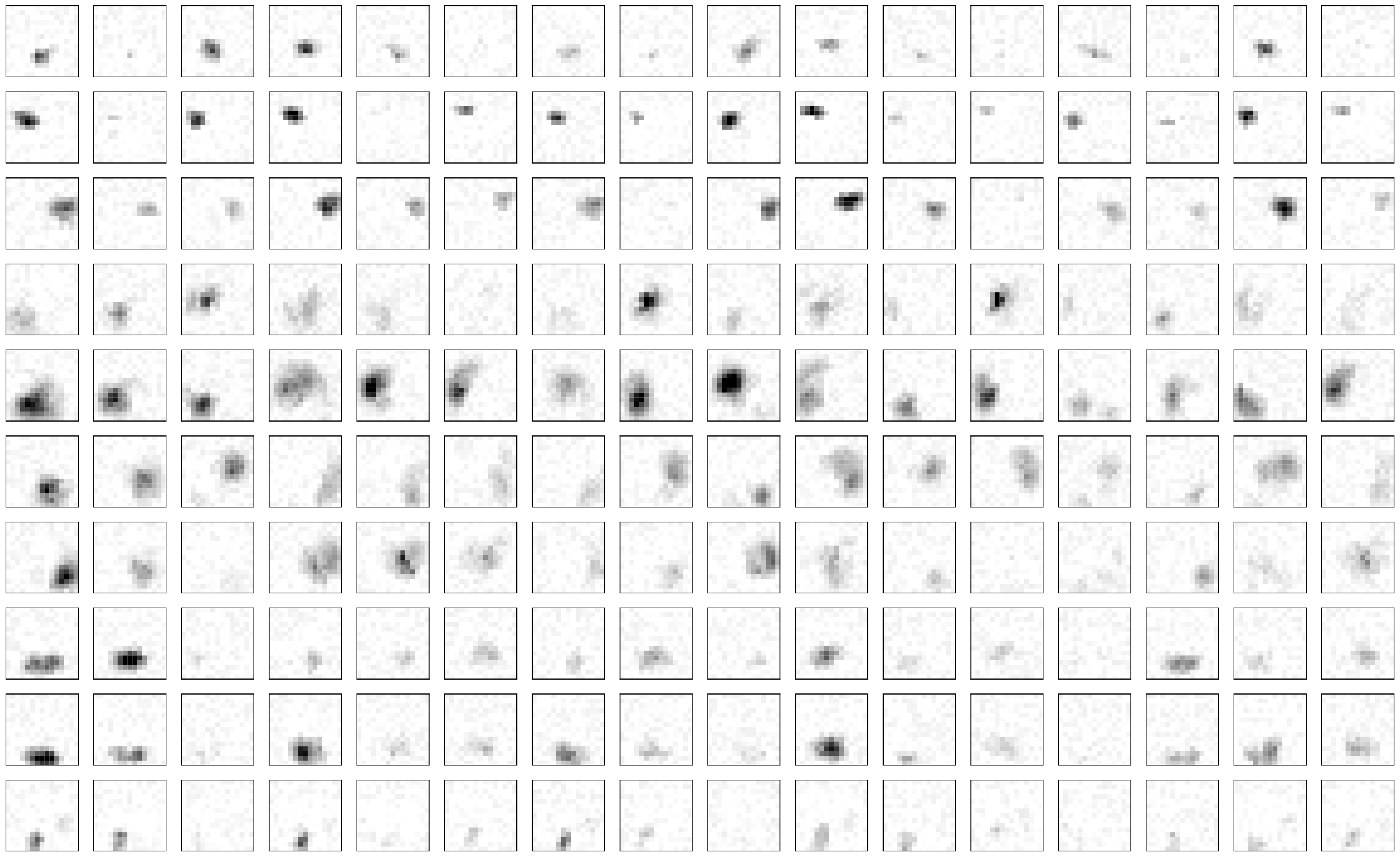}
         \caption{Synaptic delays}
     \end{subfigure}
     \hfill
        \caption{Features learnt by the connections between the first and second layers. Darker color indicates stronger connections (higher weights and lower delays).}
        \label{fig:features_l2}
\end{figure}

\begin{figure}
     \centering
     \begin{subfigure}[b]{0.45\textwidth}
         \centering
         \includegraphics[width=\textwidth]{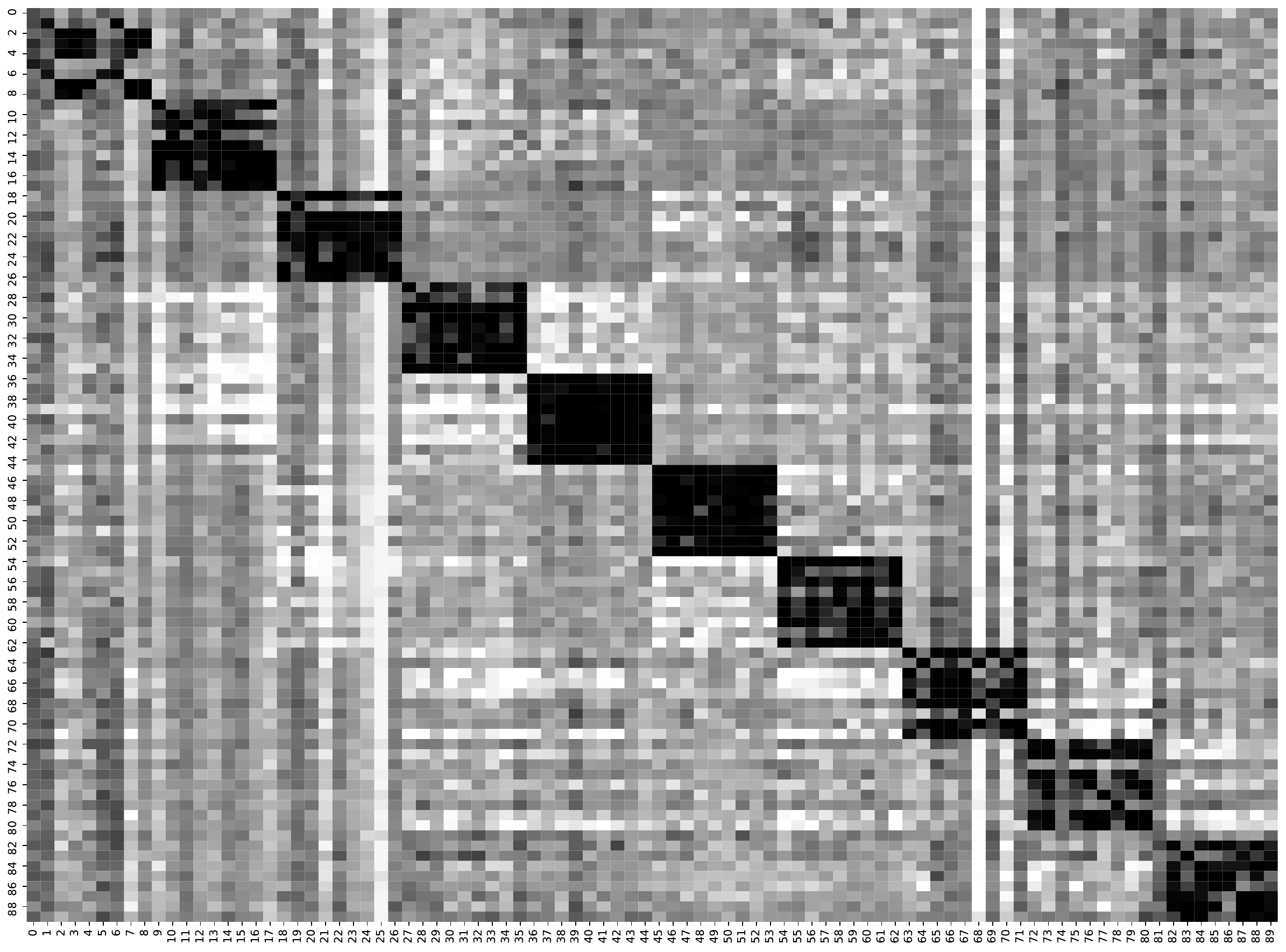}
         \caption{Synaptic weights}
     \end{subfigure}
     \hfill
     \begin{subfigure}[b]{0.45\textwidth}
         \centering
         \includegraphics[width=\textwidth]{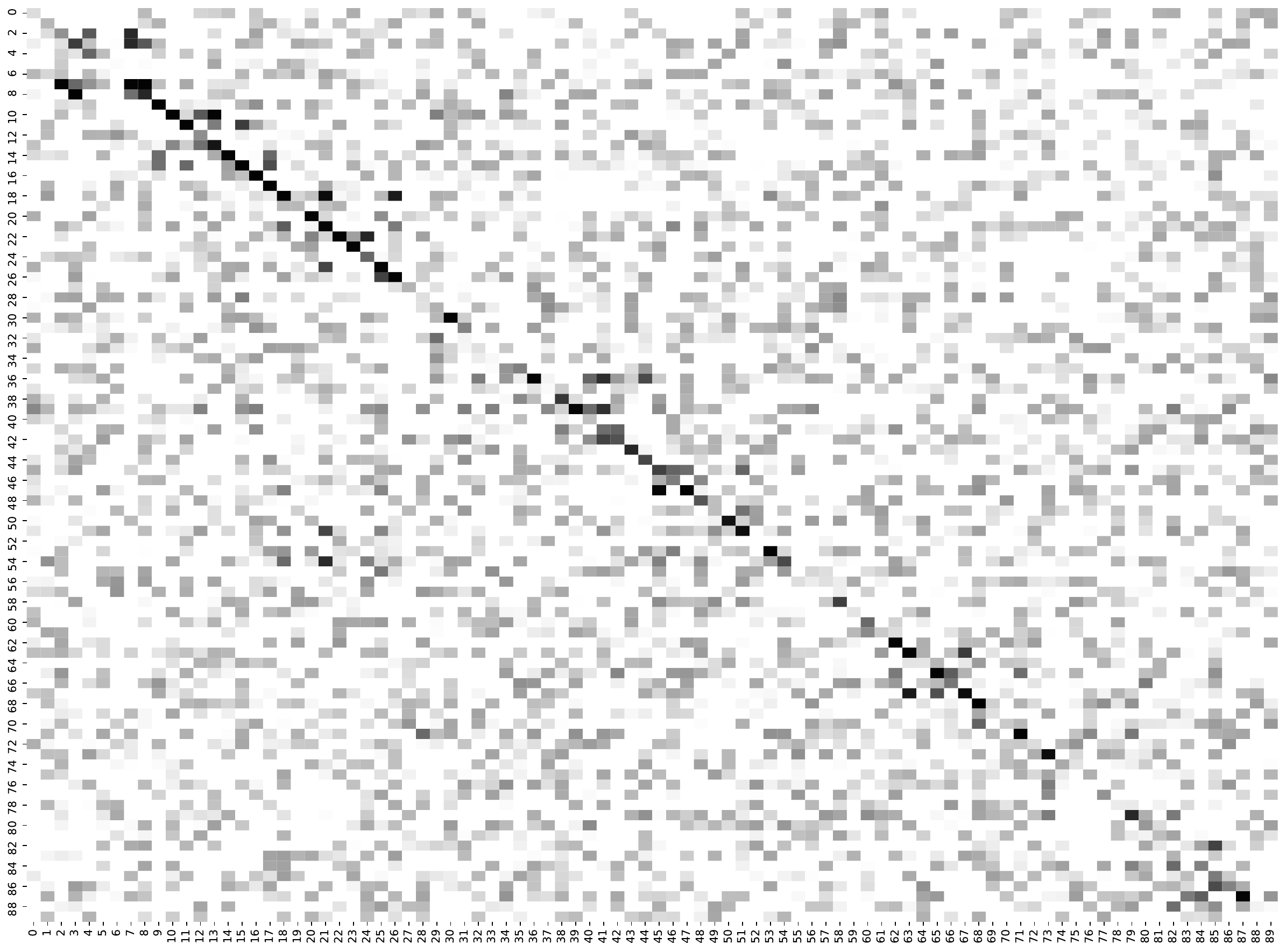}
         \caption{Synaptic delays}
     \end{subfigure}
     \hfill
        \caption{Learnt information by the lateral connections of decision layer over the training set. Darker color indicates stronger connections (higher weights and lower delays).}
        \label{fig:features_rec}
\end{figure}

Following the complete training of the first layer, we proceed to train the second layer of the model using a reinforcement approach. To train this layer, we first present each action to the first layer and then propagate its activity to the second layer. During this presentation process, whenever a decision neuron spikes, we calculate the changes in weights and delays according to R-STDP and RDL, and apply the changes at the end of the presentation when the last neuron spikes. Additionally, any changes in weights, delays, and thresholds due to the other employed mechanisms are also applied at the end of the stimulus presentation. We continue training this layer until all the neurons reach to the frozen state. Fig. \ref{fig:features_l2} and Fig. \ref{fig:features_rec} illustrate the learned features in the second layer and the recurrent synapses of the decision layer, respectively.

\subsection{Evaluation phase}

During the inference phase, the model operates on the testing dataset using a process similar to the training phase, but without updating the weights and delays. The accuracy of the proposed model, as well as other models, on the DVS128-Gesture dataset are provided in Table \ref{tab:acc}.


\begin{table}[!h]
\centering

\scalebox{0.75}{
    \begin{tabular}{| c | c | c |}
    \hline
    Category & Model & Accuracy(\%) \\
    \hline
    & & \\
    \multirow{2}{20em}{\centering{Deep Neural Network Models}}
    
    & Wang et al. (2019) \cite{1.1} & 97.08 \\ 
    & Innocenti et al. (2020) \cite{1.2} & 99.58 \\
    & & \\
    \hline
    & & \\
    \multirow{3}{20em}{\centering{Spiking Neural Network with Supervised Learning}}
    
    & Shrestha et al. (2018) \cite{2.1} & 93.64 \\
    & Kaiser et al. (2018) \cite{2.2} & 95.54 \\
    & Kaiser et al. (2019) \cite{2.3} & 92.7 \\
    & & \\
    
    \hline
    & & \\
    \multirow{3}{20em}{\centering{Biologically Plausible Spiking Neural Network}}
    
    & Gruel et al. (2021) \cite{3.1} & *78.08 \\
    & Iyer et al. (2021) \cite{3.2} & 76.13 \\
    & \textbf{Our proposed Model} & \textbf{90.52} \\
    & & \\
    
    \hline
    \end{tabular}
}
\caption{Accuracy of the proposed model along with other models on the DVS128-Gesture dataset. the * model was only tested on three types of activities, and the proposed model achieved an accuracy of $98.24\%$ for these specific activities.}
\label{tab:acc}
\end{table}

The results indicates that deep neural networks perform exceptionally well in solving activity recognition problem. Moreover, spiking neural networks trained with supervised learning also demonstrate satisfactory performance. However, biologically plausible models exhibit a relatively weaker performance in activity recognition. Nevertheless, our proposed model, despite being biologically inspired, achieves competitive performance compared to other models that lack biological justification.

\subsection{Validation}

The natural question that arises is the impact of each mechanism employed in the proposed model on its performance in the action recognition task. To address this question, we conducted a series of additional experiments. In each experiment, we disabled a single mechanism and observed the resulting changes in the model's behavior. Furthermore, we evaluated the extent to which the performance of the model in the action recognition task decreased when each mechanism was disabled.

\subsubsection{Impact of Intervallic Homeostasis Regulation on Neural Activity}

\begin{figure}[!h]
\centering
\includegraphics[width=0.85\textwidth]{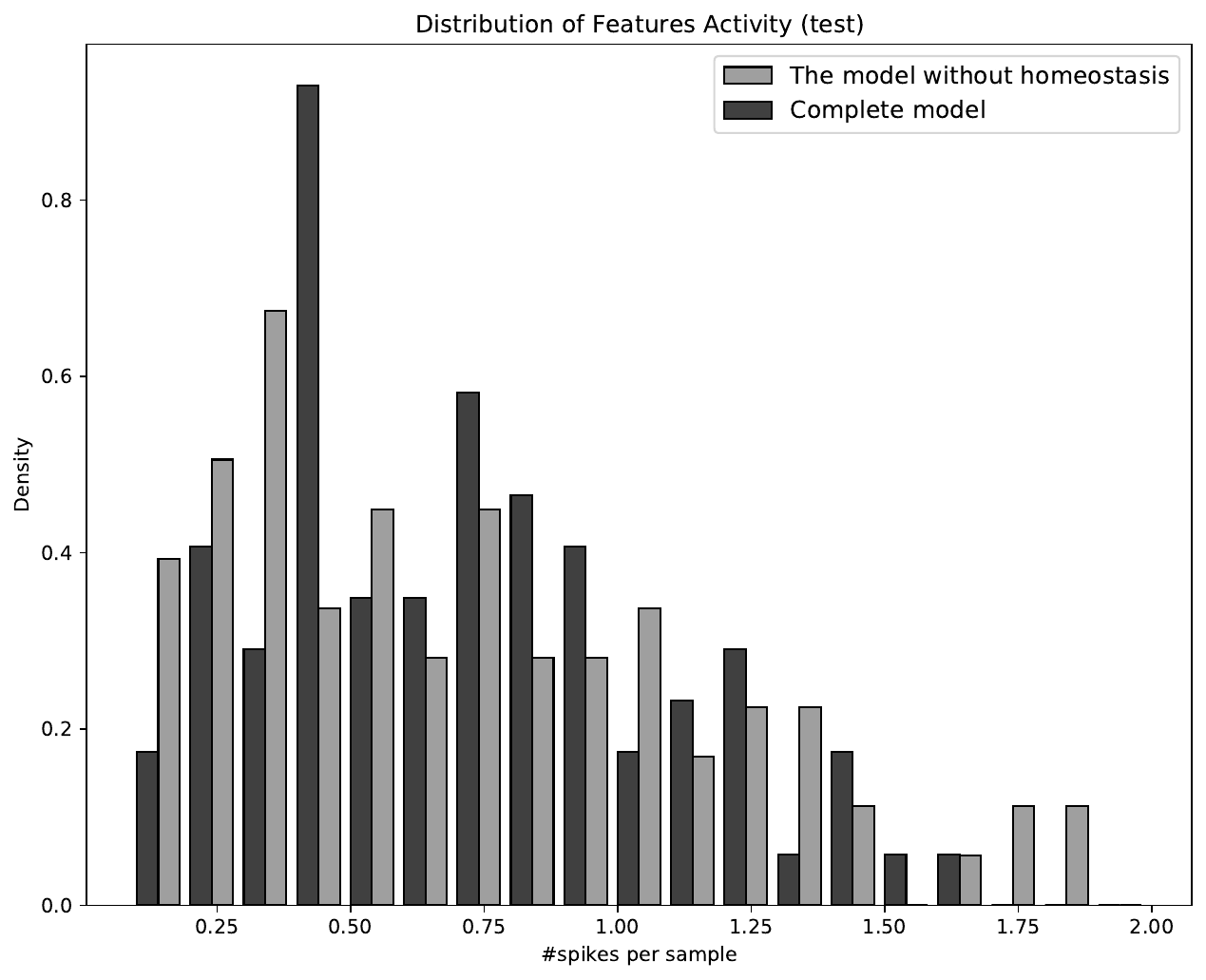}
\caption{Distribution of feature map activity levels for the complete model and the model without intervallic homeostasis regulation on the test set.}
\label{fig:homeo_res}
\end{figure}

To investigate the effect of intervallic homeostasis, we trained a second model $M_1$ similar to the proposed model M, with the only difference being the deactivation of this mechanism in the second layer. The results, depicted in Fig. \ref{fig:homeo_res}, show that by disabling the intervallic homeostasis mechanism, a substantial number of neurons exhibit zero or near-zero activity. In contrast, M maintains a small number of neurons with low activity. Moreover, in $M_1$, certain neurons display unusual high activity within the network. This imbalanced neuronal activity can impede the effective utilization of decision neurons in this model, thereby reducing its performance. For example, by evaluating $M_1$, we observed a 0.5\% and 2.1\% decrease in its accuracy, compared to M, on the train and test sets, respectively.


\subsubsection{Intervallic Threshold Adaptation Impact}

\begin{figure}[!h]
\centering
\includegraphics[height=13cm, width=\textwidth]{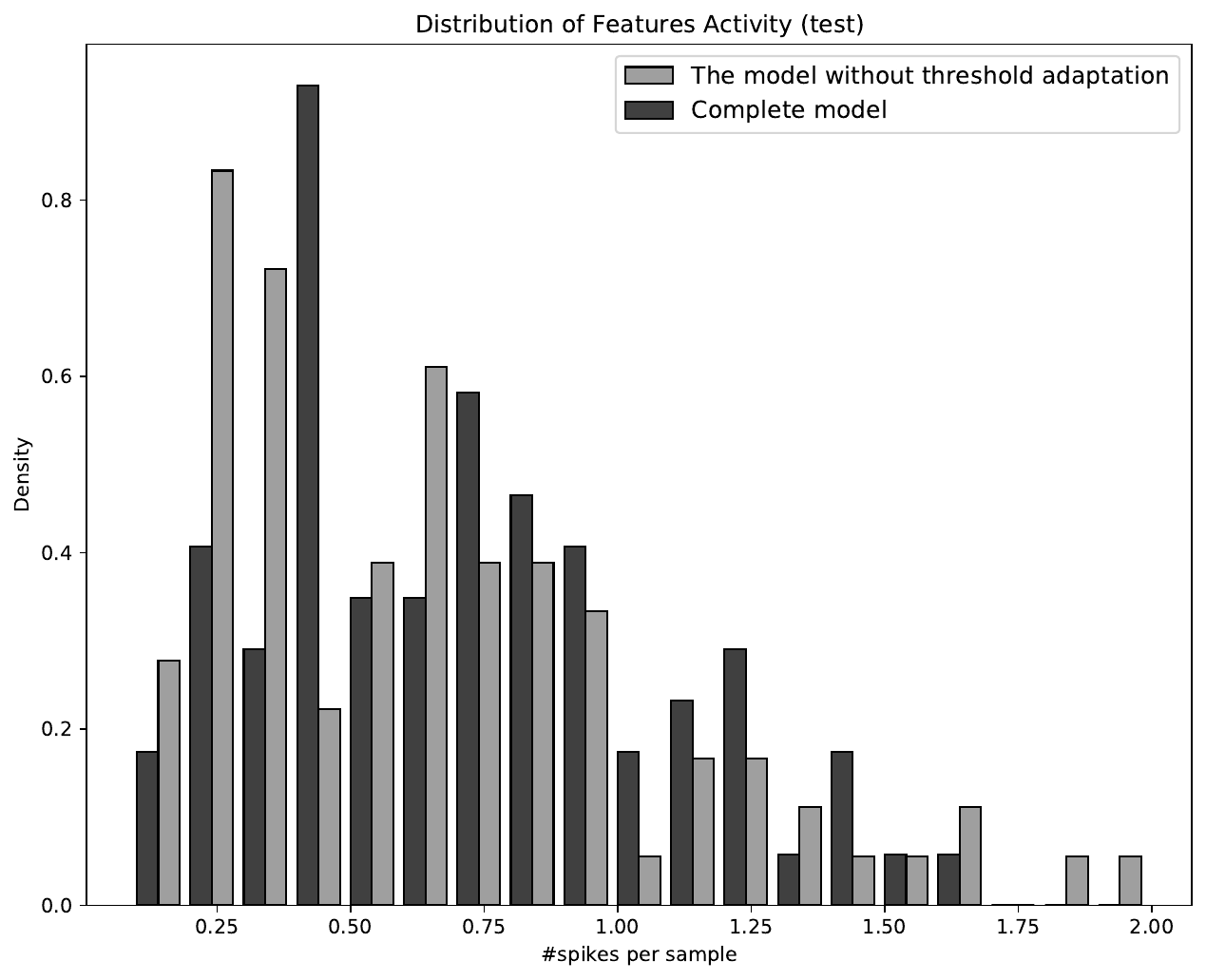}
\caption{Distribution of feature map activity levels in the experiments with and without intervallic threshold adaptation.}
\label{fig:ta_res2}
\end{figure}

Similar to the previous experiment, we deactivated the intervallic threshold adaptation mechanism in the proposed model to examine its impact. The results, illustrated in Fig. \ref{fig:ta_res2}, reveal that when this mechanism is deactivated, a significant number of neurons exhibit both very low and very high activity levels. As previously mentioned, this imbalance in neural activity can have a detrimental effect on the performance and learning capabilities of the model.

\begin{figure}[!h]
\centering
\includegraphics[height=8cm, width=\textwidth]{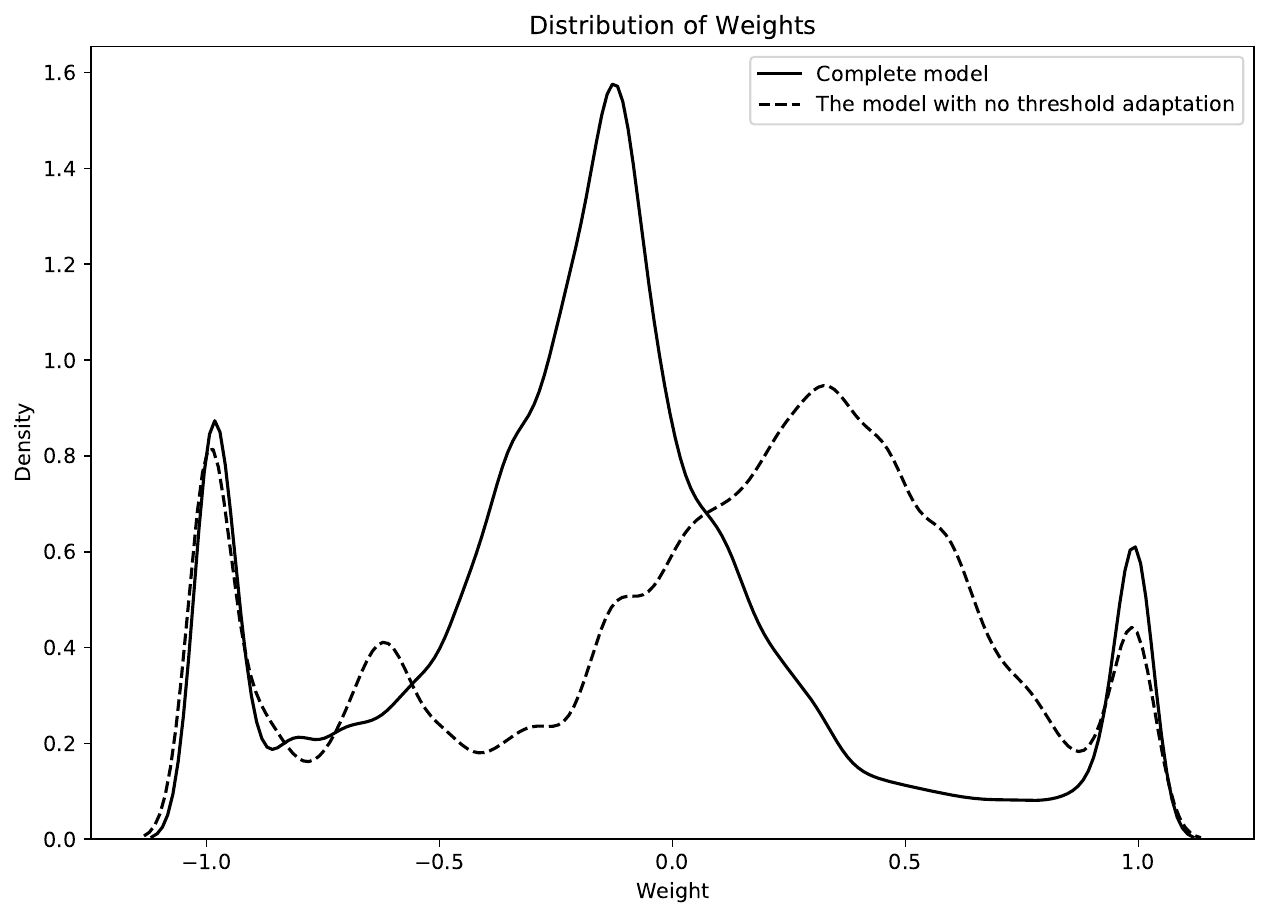}
\caption{Distribution of the synaptic weights, connecting the first layer and decision layer in in the experiments with and without intervallic threshold adaptation.}
\label{fig:ta_res1}
\end{figure}

In addition, Fig. \ref{fig:ta_res1} depicts the distribution of weights of the synapses connecting the first and second layers of the model (similar distribution can be observed for the delays as well). It is evident that the intervallic homeostasis regulation mechanism plays a crucial role in increasing the strength (absolute value) of weights to address the issue of imbalanced activity. However, this action can have an adverse effect on the learning process of the model. Specifically, weights that do not contribute to the correct decision of the model attain large values, leading to increased noise and significantly reduce the overall performance of the model. By deactivating the intervallic threshold adaptation mechanism in the proposed model, we observed 1.45\% and 3.16\% decrease in the accuracy on the train and test sets, respectively.

\subsubsection{Impact of the Modified R-STDP and RDL for Inhibitory Neurons}

\begin{figure}[!h]
\centering
\includegraphics[height=8cm, width=\textwidth]{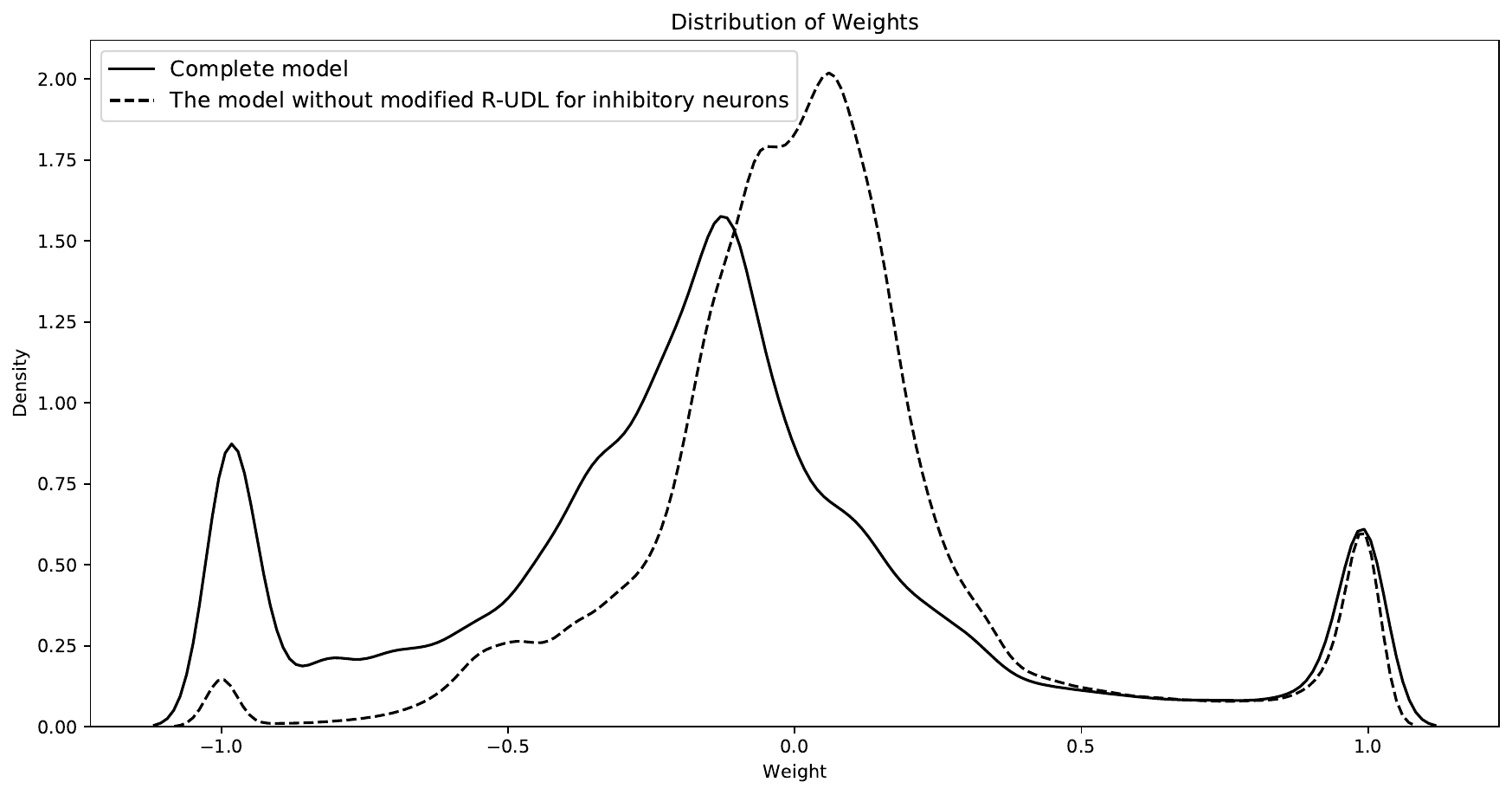}
\caption{Distribution of the synaptic weights connecting first layer to the decision layer in the models with and without modified RUDL and R-STDP. Note that the synaptic delays show the same behavior in terms of connection strength.}
\label{fig:inh_rudl_res}
\end{figure}

To adjust the synaptic delays of the pre-synaptic inhibitory neurons, we have introduced a different learning rule as provided in section \ref{sec:inh_rudl}. We investigated the impact of this mechanism on the model's performance by training a model $M_2$ where the same delay the learning rule for pre-synaptic inhibitory and excitatory neurons was utilized. Using a similar learning rule as excitatory neurons could make the pre-synaptic inhibitory neurons ineffective since the strength of synapses connected from them to other neurons would weaken as the model learns. This occurrence arises from the misassumption of R-STDP, which associates higher rewards exclusively with excitation, leading to a reduction in the strength of inhibitory synapses (the absolute value of weights moves toward zero). The final states of the models, as depicted in Fig. \ref{fig:inh_rudl_res}, verifies this deduction. Moreover, the results show that $M_2$ is 1.98\% and 1.58\% less accurate than M on train and test sets, respectively.


\subsubsection{Impact of Homeostasis on Decision}

\begin{figure}[!h]
\centering
\begin{subfigure}[b]{0.49\textwidth}
\centering
\includegraphics[width=\textwidth]{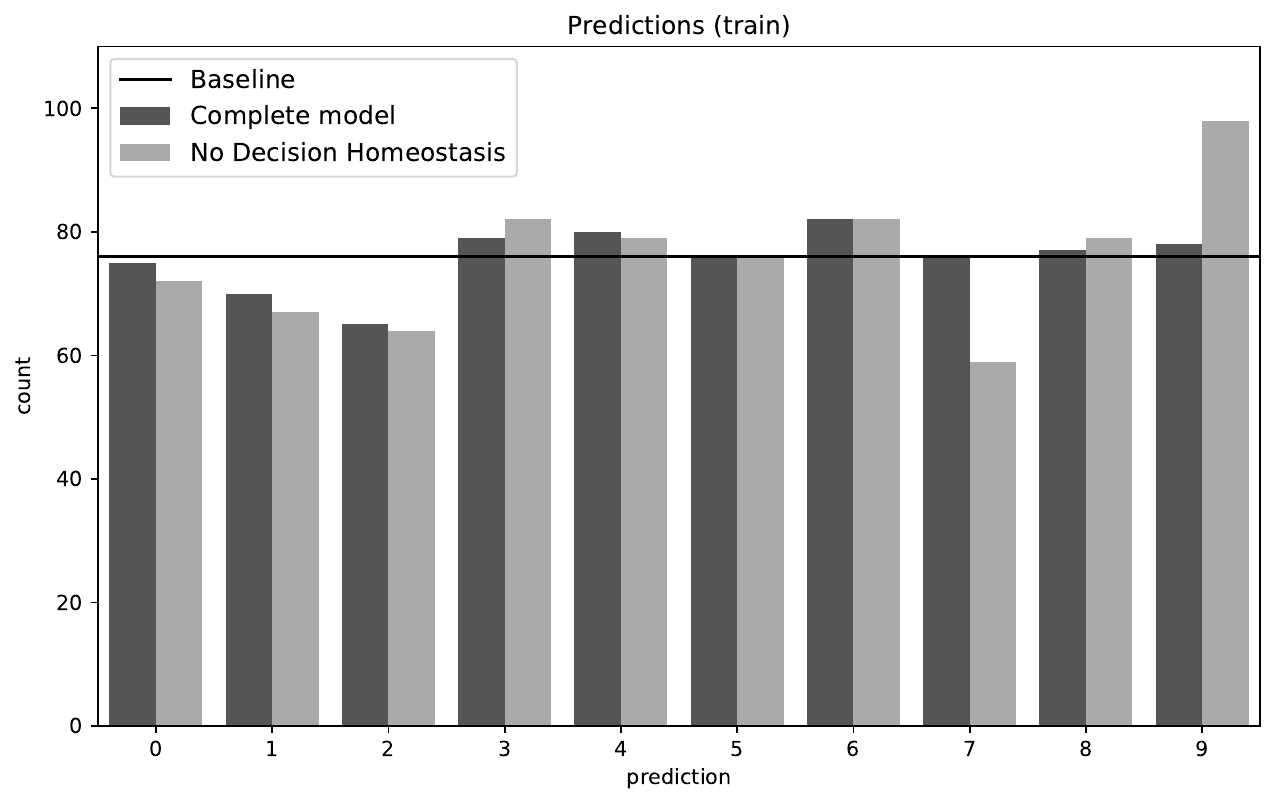}
\caption{Training dataset}
\end{subfigure}
\hfill
\begin{subfigure}[b]{0.49\textwidth}
\centering
\includegraphics[width=\textwidth]{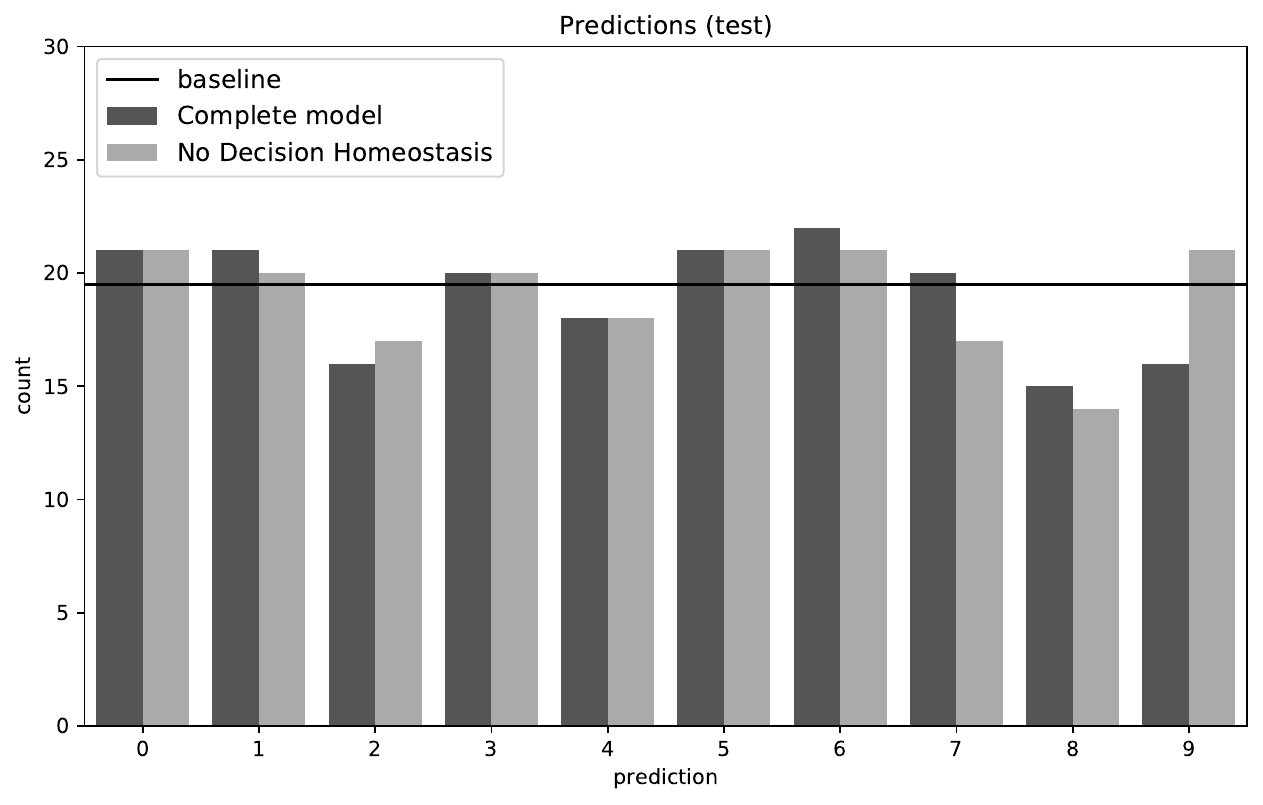}
\caption{Testing dataset}
\end{subfigure}
\hfill
\caption{Prediction of different activities by the models with and without decision homeostasis}
\label{fig:dec_homeo_res}
\end{figure}

One of the challenges encountered in the DVS128-Gesture dataset pertains to the variable distribution of spike counts within the input data. For example, some activity classes, like clockwise hand rotation exhibit a large number of active neurons, whereas there exist other classes, like playing the drums that involve only a few active neurons. This discrepancy in distribution across different classes can introduce a bias in the learning process, favoring high-activity classes due to the fact that the neurons corresponding to these classes become the winners. Moreover, the learning progress for classes with lower neuron activity typically proceeds at a slower pace under normal circumstances. To mitigate this issue, the proposed model incorporates a decision homeostasis mechanism aimed at addressing these challenges. In order to evaluate the impact of this mechanism, an alternative model $M_3$ was trained, resembling the proposed model but with the decision homeostasis mechanism disabled. The results, depicted in Fig. \ref{fig:dec_homeo_res}, reveal that the number of model decisions is closer to the target number of each activity class for both the training and testing datasets. This regulation mechanism on the decision contributes to an improvement in accuracy, with M exhibiting a 2.50\% increase on the train and a 3.68\% increase on the test set, compared to the model without homeostasis on decision, $M_3$.

\subsubsection{Decentralization Impact}
\begin{figure}
\centering
\begin{subfigure}[b]{0.49\textwidth}
\centering
\includegraphics[width=\textwidth]{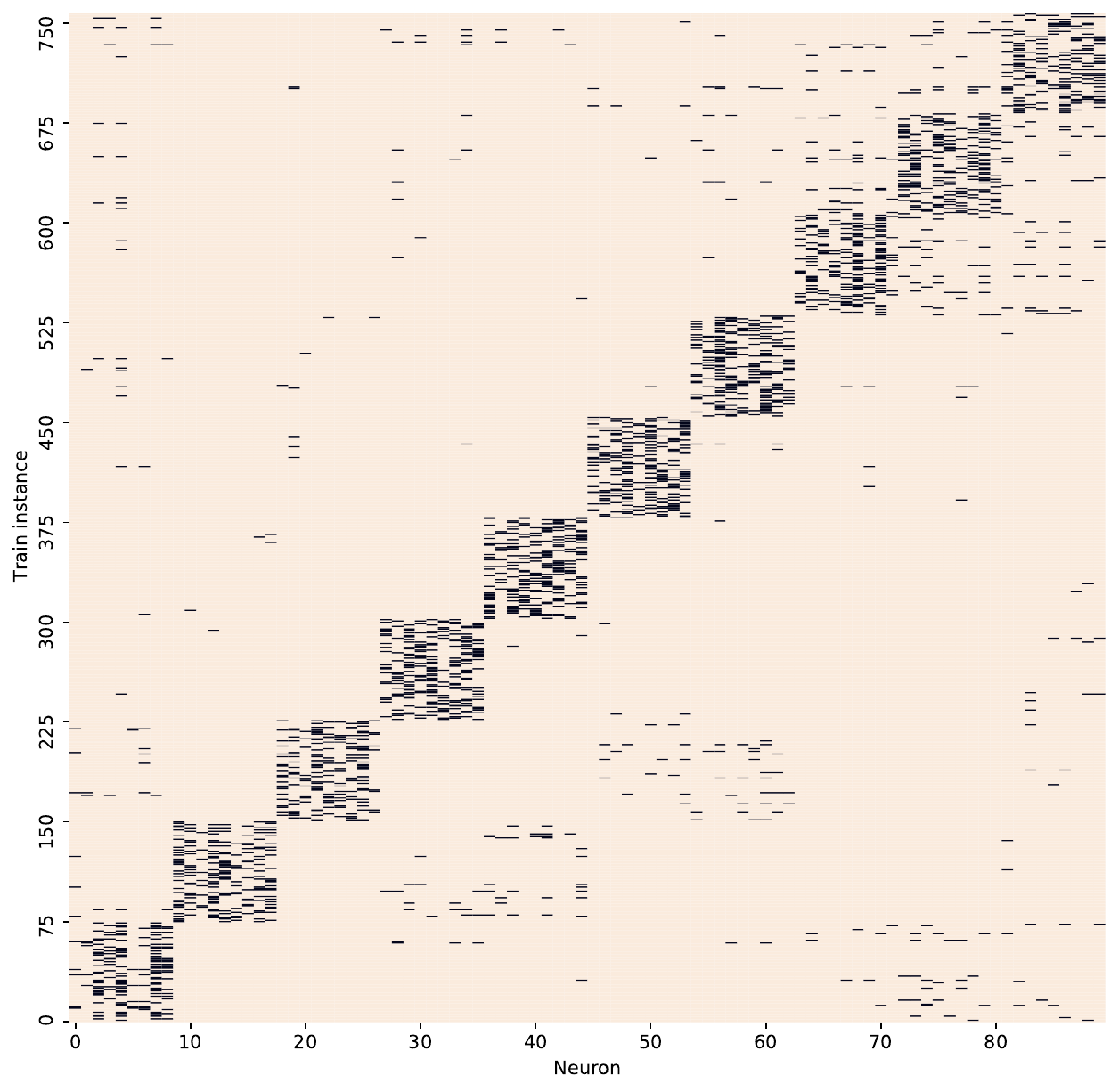}
\caption{Train set with activated decentralization}
\end{subfigure}
\hfill
\begin{subfigure}[b]{0.49\textwidth}
\centering
\includegraphics[width=\textwidth]{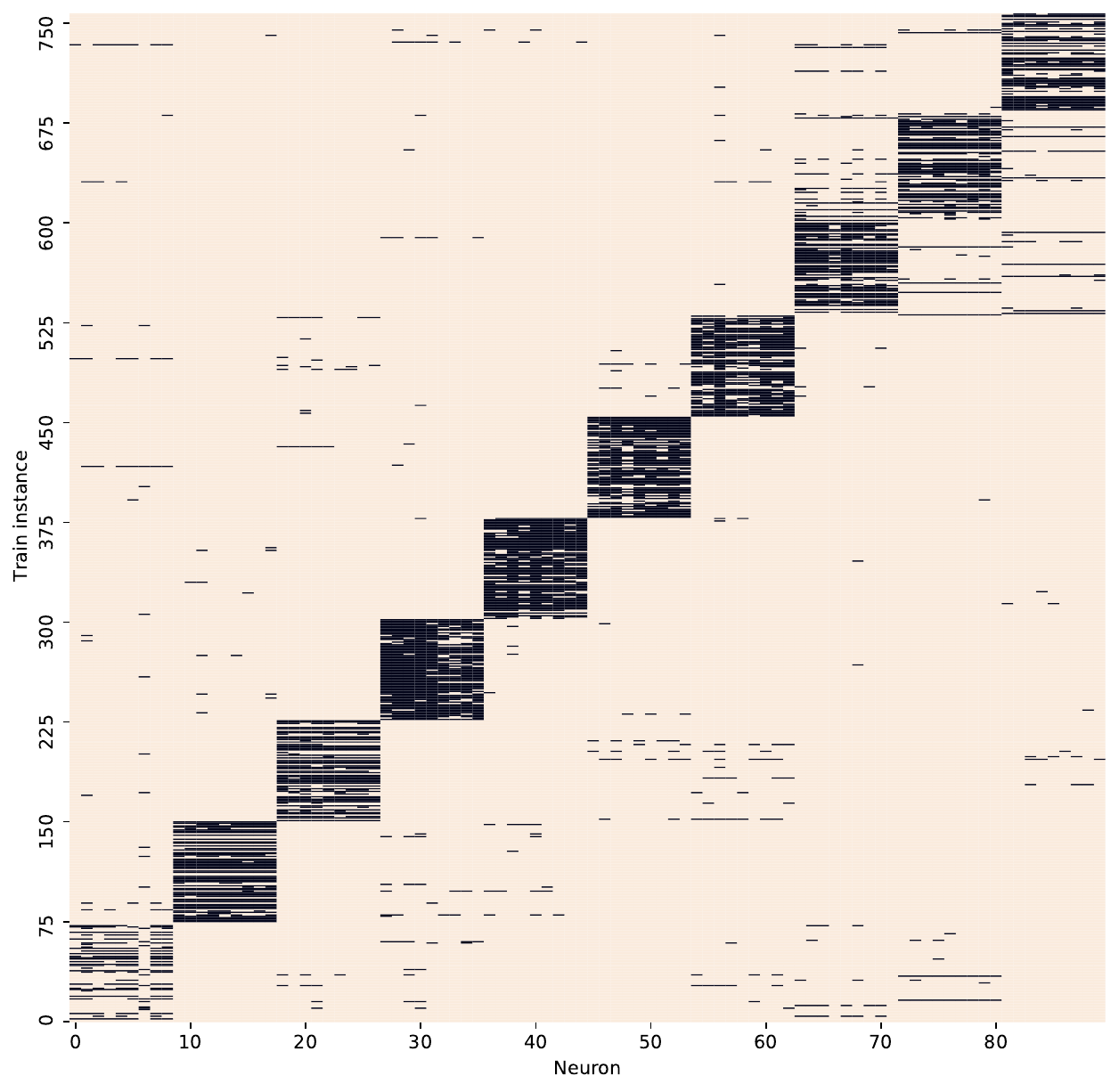}
\caption{Train set with deactivated decentralization}
\end{subfigure}
\hfill
\begin{subfigure}[b]{0.49\textwidth}
\centering
\includegraphics[width=\textwidth]{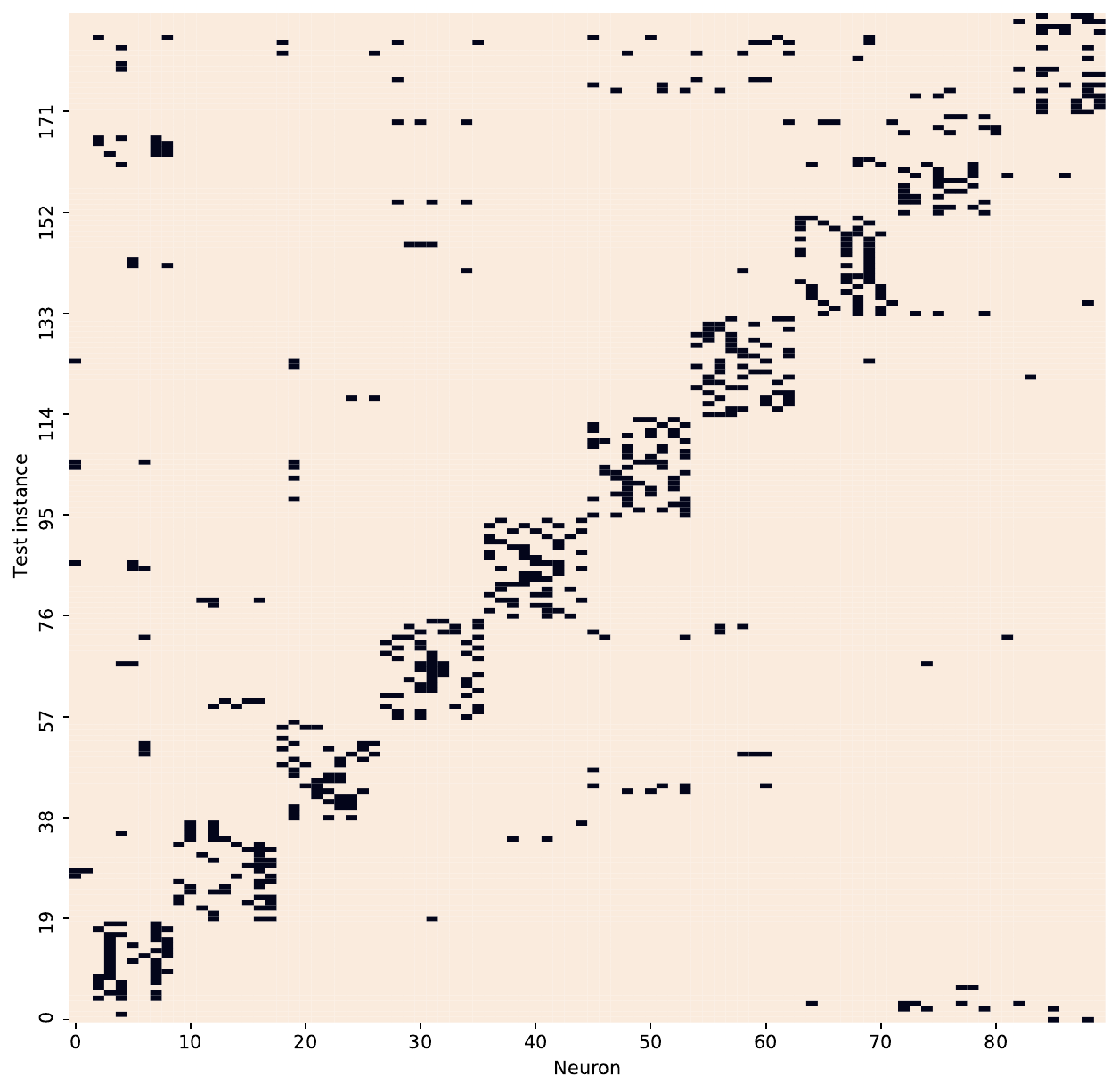}
\caption{Test set with activated decentralization}
\end{subfigure}
\hfill
\begin{subfigure}[b]{0.49\textwidth}
\centering
\includegraphics[width=\textwidth]{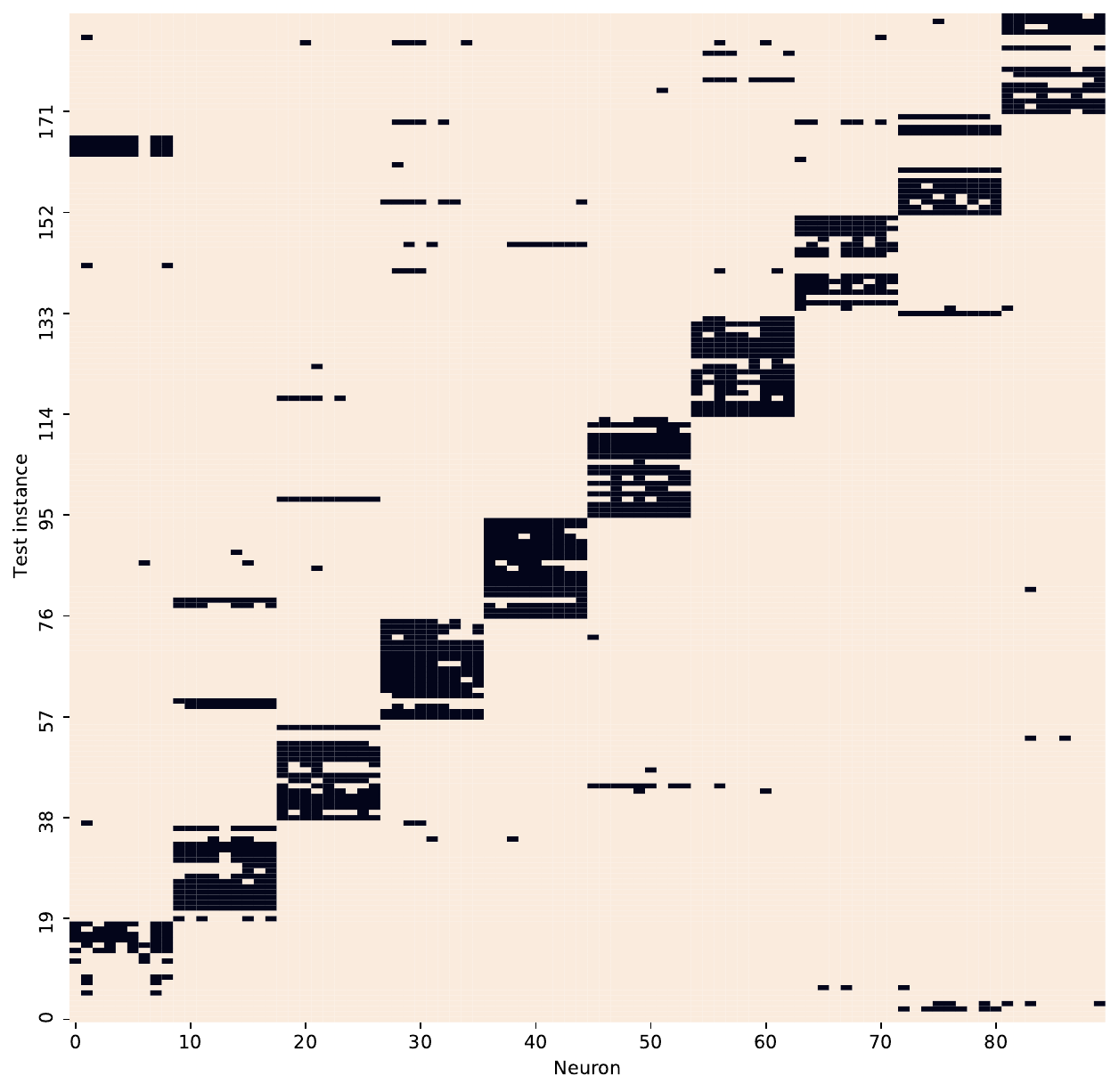}
\caption{Test set with deactivated decentralization}
\end{subfigure}
\hfill
\caption{Active decision neurons by presenting the train and test sets to the models with and without decentralization.}
\label{fig:decen_res}
\end{figure}

One of the key mechanisms in the proposed model is the decentralization. Upon deactivating this mechanism, the accuracy of the proposed model decreased by $4.88\%$ on the train and $7.39\%$ on the test set. Fig. \ref{fig:decen_res} illustrates that the primary factor causing this decrease is the ineffective allocation of neurons in learning features. For instance, when this mechanism is deactivated, all neurons associated with the presented activity class spike. In other words, all neurons learn a common set of features that are present in all samples of that activity. However, each activity class may have its own variations and unique features that are not shared among its samples. Consequently, simultaneous learning of features from all samples can disrupt neuron's ability to learn distinctive intra-class features. Moreover, as shown in the Fig. \ref{fig:decen_res},  it is evident that in numerous instances, when an activity is presented to the model, none of the neurons exhibit any spike. This indicates the insufficient resources to learn the features of samples activities that do not possess the majority. Conversely, enabling the decentralization mechanism facilitates the coordinated learning of all neurons, thereby reducing the occurrences where none of the neurons spike when an activity is presented to the model.

\subsubsection{Impact of lateral Connections}

lateral connections can have a significant impact in various scenarios, particularly in learning long activities. To assess the influence of these connections in the proposed model, similar to the other mechanisms, we trained a model without incorporating these connections, namely $M_4$. The accuracy of this model exhibited a decrease compared to $M$, with a reduction of 1.53\% on the training dataset and 1.58\% on the test dataset. 

\begin{figure}
    \centering
	\includegraphics[height=6cm, width=\textwidth]{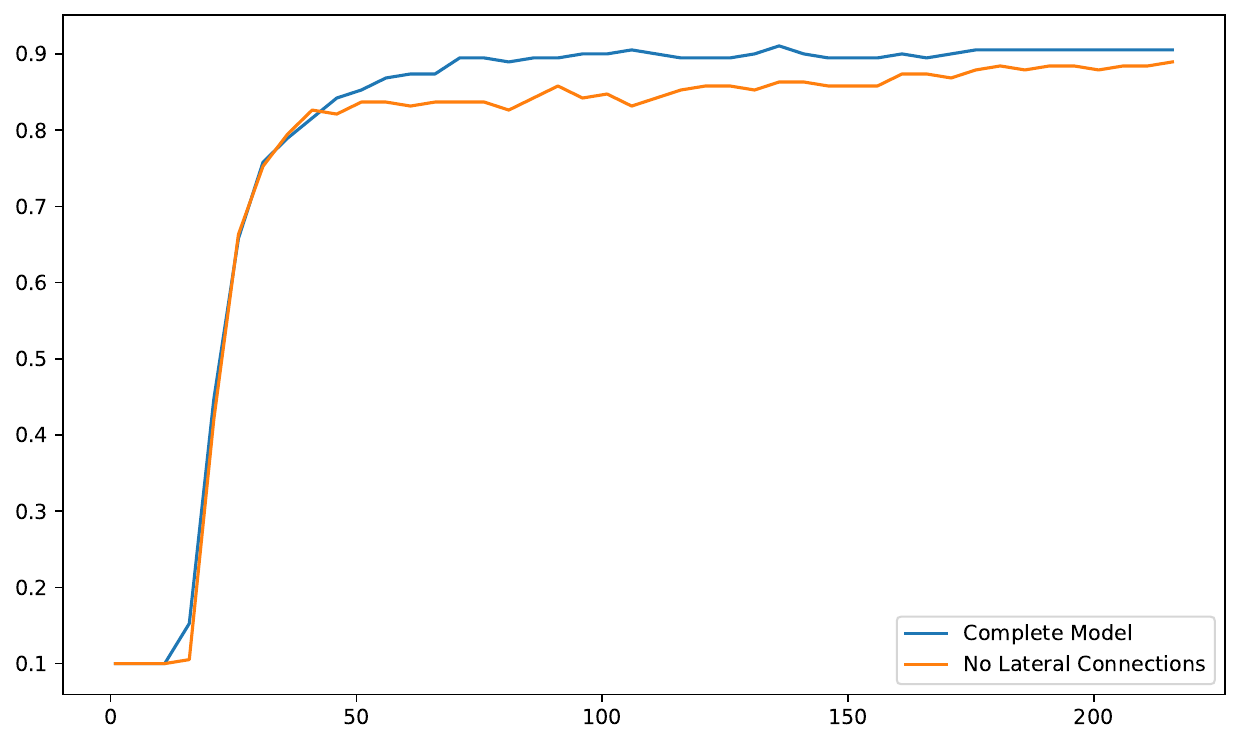}
 \caption{Accuracy of the proposed model with and without lateral connections in the case of limiting the time (number of frames) in the test set. In other words, point $x$ on the horizontal axis represents a dataset where only the initial $x$ frames of each sample are selected.}
    \label{fig:rec_res}
\end{figure}

The results presented in Fig. \ref{fig:rec_res} reveal a noticeable performance gap between the model without recurrent connections and the complete model as the number of input frames increases. This disparity can be attributed to the advantage provided by recurrent connections in capturing long-term features.

\subsubsection{Impact of Delays in Decision Layer}

Learning the temporal patterns in the proposed model is performed by three model components: UDL in the first layer, RDL in the decision layer, and lateral connections. In this section, we will examine the impact of reinforcement delay learning in the decision layer. Therefore, besides the main model, M, we trained two additional models. The first model, $M_5$, had fixed delay values for all synapses in the decision layer, and these delays remained unchanged during the training. The second one, $M_6$, had random delay values assigned to all synapses, and these delays were also not modified during training. Upon evaluating the accuracy of these two modified models, it was observed that the model with fixed delays experienced a decrease in accuracy of 1.85\% on the train set and 3.15\% on the test dataset, compared to M. Similarly, the model with random delays exhibited a 3.3\% decrease in accuracy on the train set and 3.68\% on the test set, compared to M. These findings highlight the importance of RDL in the decision layer, as omitting or randomly assigning delays adversely affects the model's performance.

\section*{Conclusion}
The brain, being an extensively intricate and enigmatic structure, performs various complex functions and effectively solves challenging problems. One of these problems is recognizing activities, which the brain accomplishes efficiently, quickly, and with minimal energy consumption. Therefore, gaining insights into how the brain tackles this problem can contribute to the development of more effective models. Unfortunately, numerous mechanisms and functions of the brain remain elusive, leading to a lack of comprehensive understanding. Consequently, many computational models inspired by the brain often avoid incorporating its underlying mechanisms, resulting in poor performance when applied to complex tasks like activity recognition. The objective of this work was to highlight the benefits of bridging this gap in order to enhance the effectiveness of bio-plausible computational models.

In this article, we first introduced several computational processes for simulating the underlying mechanisms of the brain that are effective in understanding spatio-temporal patterns and thus effective in solving the activity recognition problem. Then, to evaluate the performance of our proposed model, we used the DVS128-Gesture dataset, a widely recognized benchmark dataset for activity recognition. Results show that the proposed model in this research, besides the alignment with biology, achieves competitive performance with highly complex models like deep neural networks with numerous layers. Finally, we showed the necessity and impact of each proposed mechanism on the performance and behavior of the proposed model.

\vfill


\begin{thebibliography}{99} 
\bibitem{mcculloch}
\newblock McCulloch, W. S. and Pitts, W. 
\newblock A logical calculus of the ideas immanent in nervous activity.
\newblock{\em The bulletin of mathematical biophysics}, 5(4):115–133.

\bibitem{1.1}
\newblock Q. Wang, Y. Zhang, J. Yuan and Y. Lu.
\newblock  Space-Time Event Clouds for Gesture Recognition: From RGB Cameras to Event Cameras.
\newblock{\em 2019 IEEE Winter Conference on Applications of Computer Vision (WACV), 2019, pp. 1826-1835, doi: 10.1109/WACV.2019.00199.}

\bibitem{1.2}
\newblock S. Undri Innocenti, F. Becattini, F. Pernici, A. Del Bimbo.
\newblock  Temporal Binary Representation for Event-Based Action Recognition
\newblock{\em 2020 25th International Conference on Pattern Recognition (ICPR), 10426-10432}

\bibitem{2.1}
\newblock Shrestha, S. and G. Orchard.
\newblock  SLAYER: Spike Layer Error Reassignment in Time.
\newblock{\em  NeurIPS (2018).}

\bibitem{2.2}
\newblock Kaiser J, Mostafa H and Neftci E. 
\newblock  Synaptic Plasticity Dynamics for Deep Continuous Local Learning (DECOLLE).
\newblock{\em Front. Neurosci. 14:424. doi: 10.3389/fnins.2020.00424}

\bibitem{2.3}
\newblock J. Kaiser et al.
\newblock  Embodied Neuromorphic Vision with Continuous Random Backpropagation. 
\newblock{\em 2020 8th IEEE RAS/EMBS International Conference for Biomedical Robotics and Biomechatronics (BioRob), 2020, pp. 1202-1209, doi: 10.1109/BioRob49111.2020.9224330.}


\bibitem{3.1}
\newblock A. Gruel and J. Martinet.
\newblock  Bio-inspired visual attention for silicon retinas based on spiking neural networks applied to pattern classification. 
\newblock{\em 	arXiv:2105.14753.}

\bibitem{3.2}
\newblock Iyer LR, Chua Y and Li H.
\newblock   Is Neuromorphic MNIST Neuromorphic? Analyzing the Discriminative Power of Neuromorphic Datasets in the Time Domain. 
\newblock{\em 	Front. Neurosci. 15:608567. doi: 10.3389/fnins.2021.608567.}

\bibitem{main_1}
\newblock Dongliang He, Zhichao Zhou, Chuang Gan, Fu Li, Xiao Liu, Yandong Li, Limin Wang, Shilei Wen. 
\newblock   StNet: Local and Global Spatial-Temporal Modeling for Action Recognition.
\newblock{\em 	AAAI 2019: 8401-8408.}

\bibitem{main_2}
\newblock Shu N, Gao Z, Chen X, Liu H. 
\newblock   Computational Model of Primary Visual Cortex Combining Visual Attention for Action Recognition.
\newblock{\em PLoS ONE 10(7): e0130569. doi:10.1371/journal.pone.0130569.}

\bibitem{hebb}
\newblock Hebb D.
\newblock The Organization of Behavior : A Neuro-psychological Theory.
\newblock{\em New York, Wiley and Sons. ISBN 9780471367277.}

\bibitem{stdp}
\newblock Caporale N., Dan Y.
\newblock Spike timing-dependent plasticity: a Hebbian learning rule.
\newblock{\em Annual Review of Neuroscience. 31: 25–46.}
 
 \bibitem{stdp_repeat}
 \newblock Bauer E. P.; LeDoux J. E.; Nader K.
 \newblock Fear conditioning and LTP in the lateral amygdala are sensitive to the same stimulus contingencies.
 \newblock{\em Nature Neuroscience. 4 (7): 687–688}

\bibitem{delay_change}
\newblock Stevens B.; Tanner S. ; Fields R. D.
\newblock Control of myelination by specifc patterns of neural impulses.
\newblock{\em J Neurosci 18, 9303–9311}

\bibitem{myelination}
\newblock Purves D, Augustine G, Fitzpatrick D, Katz L, LaMantia A, McNamara J, Williams S.
\newblock Increased conduction velocity as a result of myelination.
\newblock {\em Neuroscience. In: NeuMcNamara and S Mark Williams.}

\bibitem{myelin}
\newblock Hartline, D. K.
\newblock What is myelin?
\newblock {\em Neuron Glia Biol. 2008;4(2):153-163. doi:10.1017/S1740925X09990263}

\bibitem{myelin_p}
\newblock Almeida, R.G., Lyons, D.A.
\newblock On myelinated axon plasticity and neuronal circuit formation and function.
\newblock{\em J. Neurosci. : Off. J. Soc. Neurosci. 37, 10023–10034.}



\bibitem{LIF}
\newblock L.F Abbott.
\newblock Lapicque’s introduction of the integrate-and-fire model neuron.
\newblock{\em Brain Research Bulletin,Volume 50, Issues 5–6. doi: 10.1016/S0361-9230(99)00161-6}

\bibitem{neuron_models}
\newblock Gerstner W., Kistler W. K.
\newblock Spiking neuron models.
\newblock{\em Cambridge University Press}

\bibitem{homeo}
\newblock Turrigiano G.
\newblock Homeostatic synaptic plasticity: local and global mechanisms for stabilizing neuronal function.
\newblock {\em Cold Spring Harb Perspect Biol 4:a005736.}

\bibitem{delay_learning}
\newblock A. Nadafian and M. Ganjtabesh.
\newblock Bio-plausible Unsupervised Delay Learning for Extracting Temporal Features in Spiking Neural Networks.
\newblock {\em 	arXiv:2011.09380.}


\bibitem{myelin_piano}
\newblock Bengtsson SL, Nagy Z, Skare S, Forsman L, Forssberg H, Ullén F. 
\newblock Extensive piano practicing has regionally specific effects on white matter development.
\newblock{\em  Nat. Neurosci. 2005;8:1148–1150.}

\bibitem{rl_myelin}
\newblock Joseph B, Dyer CA.
\newblock  Relationship between myelin production and dopamine synthesis in the PKU mouse brain.
\newblock{\em  J Neurochem. 2003 Aug;86(3):615-26. doi: 10.1046/j.1471-4159.2003.01887.x. PMID: 12859675.}

\bibitem{inh_stdp}
\newblock Vogels TP, Froemke RC, Doyon N, Gilson M, Haas JS, Liu R, Maffei A, Miller P, Wierenga CJ, Woodin MA, Zenke F and Sprekeler H.
\newblock   Inhibitory synaptic plasticity: spike timing-dependence and putative network function.
\newblock{\em   Front. Neural Circuits 7:119. doi: 10.3389/fncir.2013.00119.}

\bibitem{rstdp_m}
\newblock Mozafari, M., Ganjtabesh, M., Nowzari-Dalini, A., Thorpe, S. J., and Masquelier, T.
\newblock  Bio-Inspired Digit Recognition Using Reward-Modulated Spike-Timing-Dependent Plasticity in Deep Convolutional Networks.
\newblock{\em  Pattern Recognition.}

\bibitem{polychron}
\newblock Izhikevich, E. M. (2006). 
\newblock Polychronization: computation with spikes.
\newblock{\em  Neural Comput. 18, 245–282. doi: 10.1162/089976606775093882}

\bibitem{dvs}
\newblock A. Amir, B. Taba, D. Berg, T. Melano, J. McKinstry, C. Di Nolfo, T. Nayak, A. Andreopoulos, G. Garreau, M. Mendoza, J. Kusnitz, M. Debole, S. Esser, T. Delbruck, M. Flickner, and D. Modha.
\newblock  A Low Power, Fully Event-Based Gesture Recognition System.
\newblock{\em 2017 IEEE Conference on Computer Vision and Pattern Recognition (CVPR), Honolulu, HI, 2017.}

\end{thebibliography}
\end{document}